\documentclass[10pt,twocolumn,letterpaper]{article}
\usepackage{abstract}
\usepackage{iccv}
\usepackage{times}
\usepackage{adjustbox}
\usepackage{amsmath}
\usepackage{amssymb}
\usepackage{booktabs}
\usepackage{caption}
\usepackage{graphicx}
\usepackage{multirow}
\usepackage{wrapfig}
\usepackage{xcolor}
\usepackage{xr}
\usepackage[breaklinks=true,bookmarks=false]{hyperref}
\usepackage{cleveref}
\usepackage[super]{nth}
\usepackage{authblk}
\usepackage{amsmath,amsfonts,bm}
\usepackage{multirow}
\usepackage{tabularx}
\usepackage{graphicx}
\usepackage{adjustbox}
\usepackage{amsthm}

\renewcommand{\arraystretch}{1.3}
\newcommand{\mset}[1]{\left\{\kern-.5em\left\{ #1 \right\}\kern-.5em\right\}}
\newcommand{\mmset}[1]{\{\kern-.4em\{ #1 \}\kern-.4em\}}

\newcommand{\norm}[1]{\left\Vert#1\right\Vert}

\newcommand{\abs}[1]{\left\vert#1\right\vert}

\newcommand{\set}[1]{\left\{#1\right\}}
\newcommand{\parr}[1]{\left (#1\right )}
\newcommand{\brac}[1]{\left [#1\right ]}
\newcommand{\ip}[1]{\left \langle #1 \right \rangle }
\newcommand{\Real}{\mathbb R}

\newcommand{\eps}{\varepsilon}
\newcommand{\too}{\rightarrow}

\newcommand{\loss}{\mathrm{loss}}%_{\text{ur}}}
 \makeatletter
\newtheorem*{rep@theorem}{\rep@title}
\newcommand{\newreptheorem}[2]{%
\newenvironment{rep#1}[1]{%
 \def\rep@title{#2 \ref{##1}}%
 \begin{rep@theorem}}%
 {\end{rep@theorem}}}
\makeatother

\newtheorem{theorem}{Theorem}
\newreptheorem{theorem}{Theorem}

\newreptheorem{lemma}{Lemma}

\def\eqref#1{equation~\ref{#1}}
\def\1{\bm{1}}

\def\eps{{\epsilon}}

\def\veta{{\bm{\eta}}}

\def\va{{\bm{a}}}
\def\vb{{\bm{b}}}

\def\vn{{\bm{n}}}

\def\vp{{\bm{p}}}
\def\vq{{\bm{q}}}

\def\vu{{\bm{u}}}
\def\vv{{\bm{v}}}
\def\vw{{\bm{w}}}
\def\vx{{\bm{x}}}

\def\vy{{\bm{y}}}
\def\vz{{\bm{z}}}
\def\vec1{{\bm{1}}}

\def\mA{{\bm{A}}}

\def\mI{{\bm{I}}}

\def\mP{{\bm{P}}}

\DeclareMathAlphabet{\mathsfit}{\encodingdefault}{\sfdefault}{m}{sl}
\SetMathAlphabet{\mathsfit}{bold}{\encodingdefault}{\sfdefault}{bx}{n}

\def\gD{{\mathcal{D}}}

\def\gN{{\mathcal{N}}}

\def\gS{{\mathcal{S}}}

\def\gX{{\mathcal{X}}}

\def\gZ{{\mathcal{Z}}}

\newcommand{\dist}{\textrm{d}} %distance function
\newcommand{\E}{\mathbb{E}}

\makeatletter
\renewcommand\AB@affilsepx{, \protect\Affilfont}
\makeatother

\newif\ifdark
\IfFileExists{/Users/vedaldi/.bash_profile}{
\immediate\write18{%
if defaults read -g AppleInterfaceStyle 2>/dev/null;
then echo \\darktrue > /tmp/displaymode.tex;
else echo \\darkfalse > /tmp/displaymode.tex; fi}
\input{/tmp/displaymode.tex}
\ifdark
\definecolor{pcolor}{HTML}{1E1E1E}
\definecolor{tcolor}{HTML}{C5C5C5}
\else
\definecolor{pcolor}{HTML}{FDF6E3}
\definecolor{tcolor}{HTML}{333333}
\fi
\pagecolor{pcolor}
\color{tcolor}
\hbadness=\maxdimen
\vbadness=\maxdimen
\vfuzz=30pt
\hfuzz=30pt
}{}

\graphicspath{{./},{../}}

\iccvfinalcopy
\def\iccvPaperID{4300}
\ificcvfinal\fi

\title{Augmenting Implicit Neural Shape
Representations\\ with Explicit Deformation Fields }

\author[1]{Matan Atzmon \thanks{Work partially done during an internship at Facebook AI Research.} }
\author[2]{David Novotny}
\author[2]{Andrea Vedaldi}
\author[1,2]{Yaron Lipman}
\affil[1]{Weizmann Institute of Science}
\affil[2]{Facebook AI Research}

\begin{document}

%\maketitle
\twocolumn[{%
\renewcommand\twocolumn[1][]{#1}%
\maketitle
\begin{center}
    \centering
    \vspace{-15pt}
    \includegraphics[width=\textwidth]{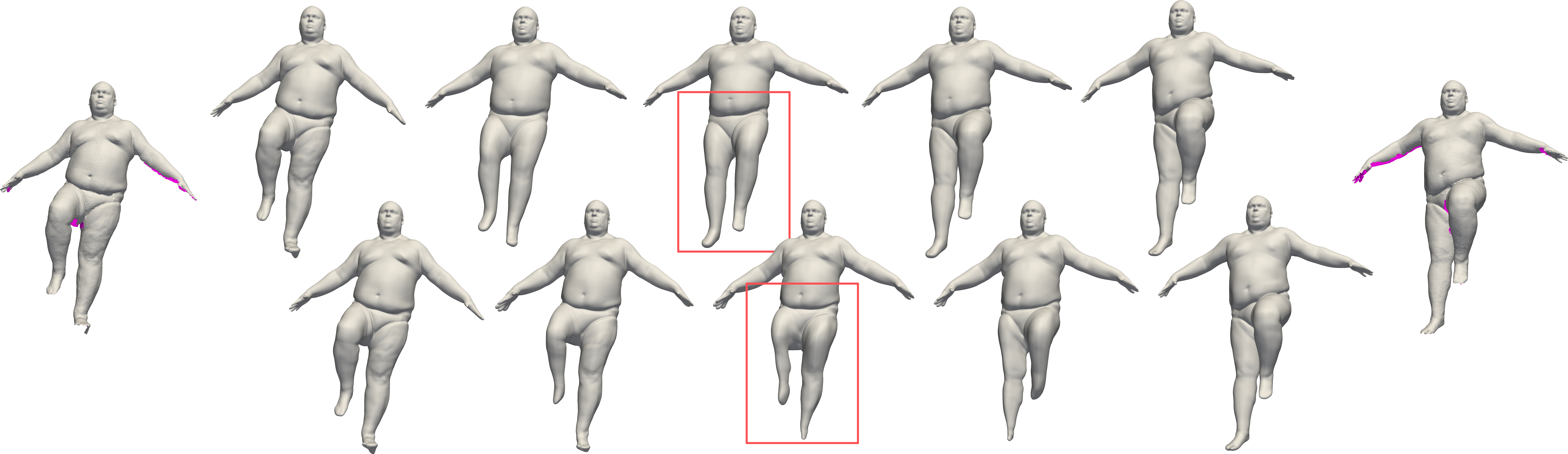}
    \captionof{figure}{Augmenting implicit neural representations with consistent deformation fields allows incorporating deformation priors, \eg, piecewise rigid, to shape space interpolations (top row); bottom row depicts state of the art implcit neural interpolation baseline; far left and right depict input raw scans.\vspace{10pt} }\label{fig:teaser}
\end{center}%
}]

\begin{abstract}\vspace{-15pt}
Implicit neural representation is a recent approach to learn shape collections as zero level-sets of neural networks, where each shape is represented by a latent code.
So far, the focus has been shape reconstruction, while shape generalization was mostly left to generic encoder-decoder or auto-decoder regularization.
In this paper we advocate deformation-aware regularization for implicit neural representations, aiming at producing plausible deformations as latent code changes.
The challenge is that implicit representations do not capture correspondences between different shapes, which makes it difficult to represent and regularize their deformations.
Thus, we propose to pair the implicit representation of the shapes with an explicit, piecewise linear deformation field, learned as an auxiliary function.
We demonstrate that, by regularizing these deformation fields, we can encourage the implicit neural representation to induce natural deformations in the learned shape space, such as as-rigid-as-possible deformations.
%Testing the new regularization on a dataset of human shapes provides improved generalization properties compared to state of the art baselines.
\end{abstract}
\saythanks
\section{Introduction}

The focal point of this work are implicit surfaces, defined as zero level-sets of scalar functions $f:\Real^3\too\Real$, 
$$
\gS=\set{\vx\in\Real^3 \ \vert \ f(\vx)=0}.
$$
The main benefit of an implicit shape representation is its ability to capture detailed
% represent high fidelity 
continuous surfaces with arbitrary topology without requiring discretization, complex data structures, or explicit sampling.

Implicit surfaces recently found good pairing with neural networks.
Representing an implicit function such as $f$ with a neural network allows to leverage its expressive power 
%(universality) and ability to adjust their degrees of freedom as non-linear function spaces 
to model accurately complex 3D scenes and objects~\cite{chibane2020neural,atzmon2019sal,mildenhall20nerf:,sitzmann2020implicit}.
%
%Since $f$ is a continuous function on a continuous domain, one of the main strong-points of implicit surfaces is their ability to represent geometric 3D shapes at a significantly high, possibly even infinite, spatial resolution.
%The recent resurgence of deep learning fueled the community's interest in implicit surfaces, as the ability of deep networks to universally approximate arbitrary functions allowed to accurately  model various complex 3D scenes or objects~\cite{chibane2020neural,atzmon2019sal,mildenhall20nerf:,sitzmann2020implicit}.
%
%Besides approximating surfaces of individual objects or scenes, 
Neural implicit surfaces can also represent \emph{collections} of 3D objects with different but related shapes.
This is typically operationalized by conditioning the implicit function $f$ with a latent code $\vz \in \Real^D$:
$$
\gS(\vz)=\set{\vx\in\Real^3 \ \vert \ f(\vx,\vz)=0}.
$$
By varying $\vz$, we can adjust the function's zero level-set to match different members of the shape set.
Indeed, recent works~\cite{niemeyer19occupancy,mescheder2019occupancy,Park_2019_CVPR} have demonstrated that, in this manner, it is possible to encode collections of complex 3D shapes of humans or human-made objects by low-dimensional latent code vectors.

Previous works have focused on the problem of obtaining high-fidelity reconstructions of a given set of 3D shapes form of noisy or incomplete measurements such as point clouds or triangle soups.
However, they have usually paid less attention to the shapes $\gS(\vz)$ generated by intermediate latent vectors $\vz$.
For example, given that $\gS(\vz_1)$ and $\gS(\vz_2)$ reconstruct two shapes from the input dataset, do $\gS((1-t)\vz_1 + t\vz_2)$, $t\in(0,1)$, form plausible intermediate shapes as well?
In this paper we answer this question, with an aim to not just reconstructing a given set of shapes from sparse observations, but also to generalize the given examples to obtain a full shape space.

Previous works on learning shape spaces have usually relied on generic latent-space regularizers such as VAE~\cite{mescheder2019occupancy,atzmon2019sal} or Auto-Decoders (AD)~\cite{Park_2019_CVPR} that mainly enforce smoothness of latent code transitions.
However, in many cases smoothness alone is not sufficient to learn plausible shapes.
For example, in Figure~\ref{fig:teaser} the bottom row shows implicit surfaces generated by interpolating the latent code obtained from state of the art implicit neural shape representations, demonstrating typical defects such as parts popping in and out of existence rather than deforming in a plausible manner.%\ma{Fix this comment.}

%An important property of a shape space embedding is its \emph{smoothness}.
%In order to represent meaningful analogies between the members of the shape set, it is essential that geometrically similar shapes are embedded to proximal latent codes.
%To this end, previous works have predominantly relied on the ability of deep nets to gracefully learn an optimally smooth shape space.
%However, as we empirically demonstrate in our work, state of the art shape space learners often violate the smoothness property, which is manifested by observing unrealistic shape deformations induced by changes in the latent space.
%Such observation calls for a more delicate treatment of the learning formulation.

Our main contribution is a new shape space regularization framework that encourages implicit neural shape representations to generate plausible interpolations of the training shapes.
Intuitively, we would like different shapes to be related by plausible deformations, such as the one arising from articulation of the underlying physical object.
However, differently from explicit shape representations such as meshes, implicit shapes do not provide point correspondences between deformations, which makes it difficult to apply standard geometric regularizers.

In order to address this challenge, we start from a careful analysis of the way neural level sets deform as the latent code changes.
Inspired by classic level-set tracking methods~\cite{stam2011velocity,tao2016near}, we parameterize the collection of (non-unique) explicit vector fields, called \emph{deformation fields}, that are consistent with the level-set deformation for a particular latent code change.
Then, among these deformation fields, we seek the one that minimizes a geometric loss, consequently enforcing a geometric prior on the implicit neural representation $f$.
We use in particular the as-rigid-as-possible energy, also known as Killing energy~\cite{ben2010discrete,solomon11as-killing-as-possible,slavcheva2017killingfusion,eisenberger2019divergence,eisenberger2020hamiltonian}, due to its ability to encourage natural elastic or piece-wise rigid shape deformations.

Although we opt for a specific geometric prior in this paper, an important advantage of our framework is its generality: it allows to incorporate any desirable deformation prior, which is expressible via explicit deformation fields, on the implicit neural shape space representation. 

%express first order constraints that tie the latent-code-induced motion of the level-set points to the surface normal while allowing the tangent motion.
%Importantly, these shifts within the tangent plane are parametrized by a novel auxiliary \emph{deformation field} which, given the latent code, predicts the span of possible tangent displacements for each implicit surface point.
%Crucially, this additional degree of freedom is then controlled by minimizing the killing energy, which enforces near-isometric level-set deformations - a property which, in many previous works~\cite{ben2010discrete,solomon11as-killing-as-possible,slavcheva2017killingfusion,eisenberger2019divergence,eisenberger2020hamiltonian}, has been shown to accompany natural shape deformations.

Empirically, we show that our framework, applied on the \emph{raw} D-Faust dataset~\cite{dfaust:CVPR:2017}, can learn a plausible shape space, considerably improving upon existing baselines (see \eg Figure~\ref{fig:teaser} top vs bottom row).
The power of our method is that, given unstructured and sparse data such as a triangle soup, it simultaneously achieves high-quality surface reconstructions, similar to previous work utilizing implicit shape representations, while also learning a space of plausible shapes generalizing the training samples.
We provide both quantitative and qualitative results in the experimental section.

\section{Related Work}

\subsection{Shape representations}

Since our work considers learning of a shape space, below we revise the most relevant 3D shape representations.

\paragraph{Implicit representations.}
Implicit shape representations define a 3D shape as a level-set of a scalar function on $\Real^3$ (\eqref{e:levelset}).
Several works~\cite{tatarchenko2017octree,wu2016learning,girdhar2016learning,choy20163d} have explored representing such functions with voxel grids, which are implicit functions evaluated at 3D grids.
Implicit functions on a continuous domain have been explored later~\cite{Park_2019_CVPR,xu2019disn,atzmon2019sal,atzmon2019controlling,chen2019learning,mescheder2019occupancy}.
Park et al.~\cite{Park_2019_CVPR} trained deep networks to label each 3D point with its signed distance from the nearest surface point on the boundary of a shape producing a signed distance field (SDF).
Disn~\cite{xu2019disn} improved the architecture of~\cite{Park_2019_CVPR}.
Preceding the deep learning era, SDFs were successfully applied to represent scene geometries in~\cite{newcombe2011kinectfusion}.
Atzmon et al.~\cite{atzmon2019sal} fitted signed distance fields to raw data via a sign agnostic similarity loss, and \cite{atzmon2019controlling} proposed to improve SDF learning via a differentiable parametrization of samples from the neural level-set. Deep occupancy fields, a specific kind of neural SDFs which reduce the distance function only to its sign, have been proposed in~\cite{chen2019learning,mescheder2019occupancy}.

Several works model shapes as compositions of well-defined 3D primitives.
Genova et al.~\cite{genova2019learning,genova2019deep} learn structured deep representations whose main components are Gaussian-shaped occupancy functions.
Compositions of more explicitly defined mesh primitives were introduced in~\cite{li2019supervised}. \cite{williams2020voronoinet,novotny2017learning} who split the 3D domain into Voronoi primitives and express their occupancies to delineate the surface.

The aforementioned approaches mostly improve the expressiveness of the representations to better model shape details.
As such, our contribution is complimentary, as we focus on regularizing an arbitrary neural level-set predictor in order to improve generalization to test data.

\paragraph{Explicit surfaces.}

Instead of defining a surface implicitly as a function's level-set, others explored more explicit representations, such as meshes.
Notable deep mesh predictors were introduced in~\cite{groueix20183d,wang2018pixel2mesh}.
\cite{groueix2018papier,williams2019deep} proposed a more general representation that maps points from a fixed topological space (a 2D square) to 3D euclidean space. Others~\cite{tatarchenko2016multi,richter2018matryoshka,lun20173d} represent shapes as a union over multi-view back-projections of the visible part of its surface.

\subsection{Shape interpolation}

Shape interpolation studies the problem of non-rigidly deforming a source 3D shape such that it matches a target template.
In principle, there are infinitely many possible displacement trajectories that ``flow'' one shape to the other, which calls for an efficient regularizer that selects the most natural solution.\vspace{-5pt}

\paragraph{ARAP surface regularization.}

Similar to others, our work exploits the seminal as-rigid-as-possible (ARAP) paradigm~\cite{sorkine2007rigid} that constrains deformations to be as isometric as possible.
In the context of vector fields that displace implicit surfaces, ARAP can be operationalized with the Killing energy~\cite{ben2010discrete}.
A flow field with minimal Killing energy is denoted as-Killing-as-possible vector field (KVF)~\cite{ben2010discrete}.
Solomon et al.~\cite{solomon11as-killing-as-possible} applied such KVFs to regularize 2D image warps.

KVFs were further applied to implicit surface tracking, which identifies trajectories of points attached to a time-evolving surface.
The as-killing-as-possible constraint aided a surface point tracker in~\cite{tao2016near}.
Previously,~\cite{stam2011velocity} proposed to track points with constant normal.
Similar to~\cite{stam2011velocity,tao2016near}, we tie level-set points to move consistently on the surface.
We differ from these works by using KVFs to learn the implicit surfaces' \emph{shape and motion}, rather than just tracking points on a given sequence.
This introduces several new challenges such as the requirement of an explicit parameterization of consistent vectors fields and efficient (closed form) solution for the KVF at sampled implicit surfaces.\vspace{-5pt}

\paragraph{Correspondence flows.}

Slavcheva et al.~\cite{slavcheva2017killingfusion} find a KVF that matches implicit surfaces of a pair of 3D scans.
Eisenberger et al.~\cite{eisenberger2019divergence} interpolate shapes by constructing a volume-preserving flow and minimizing a reconstruction and ARAP energies. \cite{eisenberger2020hamiltonian} improves~\cite{eisenberger2019divergence} by introducing additional momentum conservation constraints.
Among learning-based methods, Limp~\cite{cosmo2020limp} trains a mesh deformation network given ground truth 3D correspondences. \cite{jiang2020shapeflow,wang20193dn} differ from Limp by minimizing the Chamfer distance which does not require correspondence annotations.
Our method is also unsupervised and takes as input raw 3D scans. Furthermore, we are the first to jointly consider the shape interpolation in the context of implicit neural shape space learning.

\section{Method}\label{s:method}

Given a collection of input geometries, $\gX^{(i)}\subset \Real^3$, $i\in[m]:=\set{1,2,\ldots,m}$ (\eg manifold meshes, raw point-clouds, or triangle soups), our goal is to learn a \emph{shape space}.
In our case, this means a neural network $f:\Real^3\times\Real^D\too\Real$, where each latent vector $\vz\in\Real^D$ represents a shape given by the implicitly-defined surface
\begin{equation}\label{e:levelset}
\gS(\vz)=\set{\vx\in\Real^3 \ \vert \ f_\theta(\vx,\vz)=0},
\end{equation}
where $\theta\in\Real^p$ are the learnable network parameters; in the following we will drop the $\theta$ subscript of $f$ for brevity.
Two important properties of the shape space are:
(i) All input shapes are represented, \ie, for all shapes 
$
\gS(\vz^{(i)}) \approx \gX^{(i)}
$
for some $\vz^{(i)}\in\Real^d$; and
(ii) other latent vectors $\vz\in\Real^D$ correspond to \emph{plausible} shape deformations.
For example, a shape $\gS(t \vz^{(i)} + (1-t)\vz^{(j)})$ for $t\in(0,1)$ would correspond to a natural deformation between $\gS(\vz^{(i)})$ and $\gS(\vz^{(j)})$.

The first requirement of the shape spaces (i) is enforced as in previous works (detailed later).
% where, for particular latent vectors, $f$ is required to be either a Signed Distance Function (SDF) to $\gS_i$~\cite{Park_2019_CVPR,gropp2020implicit}, or an occupancy function, changing sign over $\gS_i$~\cite{mescheder2019occupancy,chen2019bsp}. We used the Sign Agnostic Loss (SAL)~\cite{atzmon2019sal} which can be used for unsigned, raw input data, and possess the desirable minimal surface property (\ie, strives to keep surface area small).
The main focus of this work is (ii), \ie, regularizing $f$ so to enforce plausible shape deformations as latent vectors change; this is done using consistent deformation fields, described next.

\subsection{Consistent deformation fields}\label{ss:consistent_fields}

Consider a smooth change in latent codes $\vz(t)$, $t\in(-\eps,\eps)$, where $\vz(0)=\vz_0\in\Real^D$ is the position at time $t=0$, and $\dot{\vz}(0)=\veta\in\Real^D$ is the speed at $t=0$.
Integrated, this process generates a curve in latent space describing a deformation of the shapes, $\gS(\vz(t))$.
We would like to incorporate a prior on this deformation.
For example, in many applications the shape deformations are  expected to be locally approximately rigid (\eg, because the deformation is driven by an articulated skeleton).

The challenge of imposing such a rigidity constraint is that the shapes are given to us implicitly as zero level-sets of a neural network, $f(\vx,\vz(t))$.
Differently from the case of meshes, where the trajectory of individual vertices is known, with implicit surfaces the motion of 3D points is only known up to a shift along the surface.
We thus propose to supplement the implicit representation with explicit consistent vector fields $\vv$ of point displacements that provide all possible explanations of the deformation of the surface $\gS(\vz(t))$, and use the latter to define deformation priors.
Consistent vector fields have been previously considered for tracking known/input sequences of implicit surfaces~\cite{stam2011velocity,tao2016near}.
Here, we use them to facilitate the learning of a space of deformable implicit surfaces.
%In contrast, we use them in higher dimension latent spaces to facilitate the learning of the implicit surfaces themselves.

We call a vector field $\vv=\vv_{\vz_0,\veta}:\Real^3\too\Real^3$ \emph{consistent} with the latent motion $\vz(t)$ if, for all $\vx\in\gS(\vz_0)$, it satisfies the constraint
\begin{equation}\label{e:consistent_def}
\frac{d}{dt}\Big\vert_{t=0}f( \vx + t \vv(\vx), \vz(t))=0.
\end{equation}
\begin{wrapfigure}[9]{r}{0.35\columnwidth}
\centering
  \vspace{-0pt}
  \hspace{-20pt}
  \includegraphics[width=0.3\columnwidth]{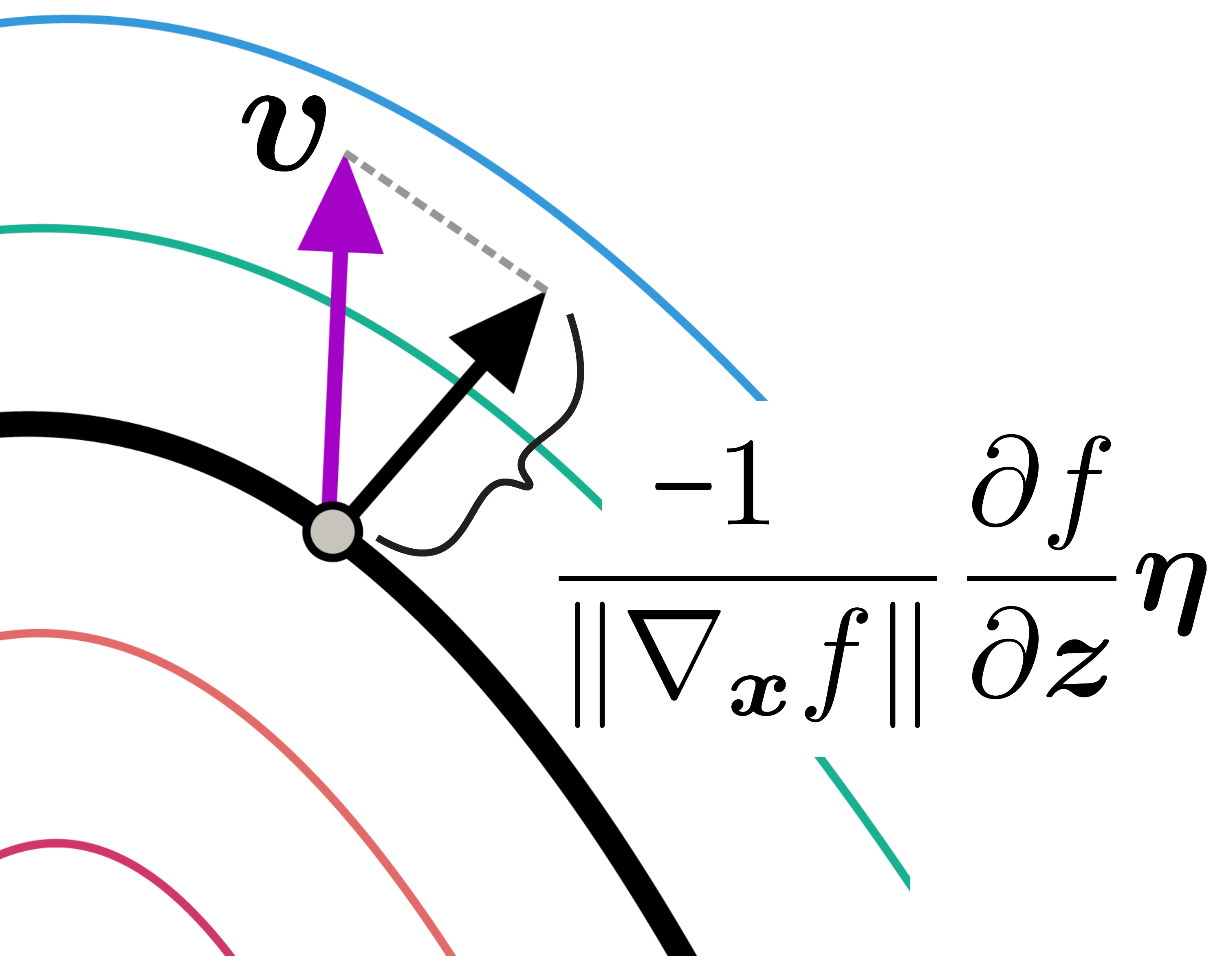}
  \captionsetup{justification=raggedright}
  \caption{ Consistent deformation field.}\label{fig:V}
\end{wrapfigure}
Intuitively, this means that, given a point $\vx$ that belongs to the implicit surface at time $0$, \ie, $\vx\in\gS(\vz_0)$, then moving $\vx$ at speed $\vv(\vx)$ will keep the point on the surface $\gS(\vz(t))$ for small (infinitesimal) times $t$.
The chain rule allows writing~\cref{e:consistent_def} more explicitly as
\begin{equation}\label{e:consistency}
\nabla_\vx f~ \vv + \frac{\partial f}{\partial \vz}\veta = 0,
\end{equation}
where $f$ is evaluated at $(\vx,\vz_0)$, and $\nabla_\vx f \in \Real^{1\times 3}$ denotes the gradient of $f$.
\Cref{e:consistency} is an under-determined linear system in $\vv\in\Real^3$ and its general solution can be written as a sum of a particular solution and the kernel of $\nabla_\vx f(\vx,\vz_0)$.
Specifically, we write $\vv = \vw +\vw^\perp$ where $\vw=\vw_{\vz_0,\veta}$ is the particular solution to~\cref{e:consistency} given by
\begin{equation}\label{e:w}
\vw = -\frac{\nabla_\vx f ^T}{\norm{\nabla_\vx f}^2}\frac{\partial f}{\partial \vz} \veta,
\end{equation}
and $\vw^\perp$ is any vector perpendicular to $\nabla_\vx f(\vx,\vz_0)$.
We let $\vw^\perp = \mP \vu$, where $\vu:\Real^3\too\Real^3$ is an arbitrary vector field, and $\mP$ is the projection matrix on $\nabla_\vx f ^\perp$ (\ie, the tangent space to $\gS(\vz_0)$ at $\vx$) defined by
\begin{equation}\label{e:P}
    \mP = \mI-\frac{\nabla_\vx f^T \nabla_\vx f}{\norm{\nabla_\vx f}^2},
\end{equation}
where $\mI\in\Real^{3\times 3}$ is the identity matrix.
In this manner, we can write the space of consistent vector fields, i.e. the fields
$$
\vv=\vv_{\vz_0,\veta,\vu}:\gS(\vz_0)\too \Real^3
$$
that satisfy \cref{e:consistency}, as
\begin{equation}\label{e:v}
\vv = \vw + \mP\vu.
\end{equation}
Intuitively, \cref{e:v} represents the space of possible motions of the surface points that are consistent with the implicit shape deformation, where the ambiguity here comes from the fact that a point $\vx$ can also move arbitrarily in the tangent directions to the surface $\gS(\vz_0)$ without compromising consistency.

\subsection{Deformation priors}

Now that consistent vector fields $\vv$ are introduced, we are ready to impose deformation priors on the shape space.
The deformation priors will take the form
\begin{equation}\label{e:loss_deform}
\text{loss}_{\text{d}}(\theta) = \min_{\vu}\E_{\vz,\veta,\vx}\, \rho(\vx ; \vv_{\vz,\veta,\vu}),
\end{equation}
where the latent code $\vz$ and variation $\veta$ are sampled according to some distribution $\gD=\gD(\Real^{D}\times\Real^{D})$, the 3D point $\vx$ is sampled from some distribution over the corresponding surface $\gS(\vz)$, and $\rho(\vx; \vv)$ is a measure of deformation induced by the field $\vv$ at location $\vx$.
As commonly done, we approximate the expectation in~\cref{e:loss_deform} with the empirical expectation
\begin{equation}\label{e:loss_emp}
  \loss_{\text{d}}(\theta) = \min_\vu \frac{1}{n}\sum_{i=1}^n \rho(\vx_i; \vv_{\vz_i,\veta_i,\vu}),
\end{equation}
where we consider $n$ samples, $(\vz_i,\veta_i)\sim \gD$ and $\vx_i\in \gS(\vz_i)$, $i\in [n]$.
Since all choices of $\vu$ correspond to consistent deformation fields, taking the minimum w.r.t.~$\vu$ is important to ensure we are considering the deformation field with lowest deformation error explaining the current shape space deformation $\vz(t)$. In the following example, we demonstrate why taking an arbitrary consistent vector field instead of the optimal one is insufficient.

\paragraph{Example.}

We provide a simple example of a rigid movement of an implicit shape with a non rigid consistent deformation field.
Consider the implicit function
$$
f(\vx,t) = \norm{\vx-t\veta}^2-1,
$$
representing a unit sphere translated with constant velocity $\veta$.
The consistency \cref{e:consistency} evaluated at $t=0$ in this case is
$$
\ip{\vx,\vv-\veta}=0
$$
for all $\vx \in \gS(0)$, \ie, $\norm{\vx}=1$.
Therefore $\vw = \frac{\ip{\vx,\veta}}{\norm{\vx}^2}\vx$
\begin{wrapfigure}[8]{r}{0.35\columnwidth}
\centering
  \vspace{-10pt}
  \hspace{-20pt}
  \includegraphics[width=0.3\columnwidth]{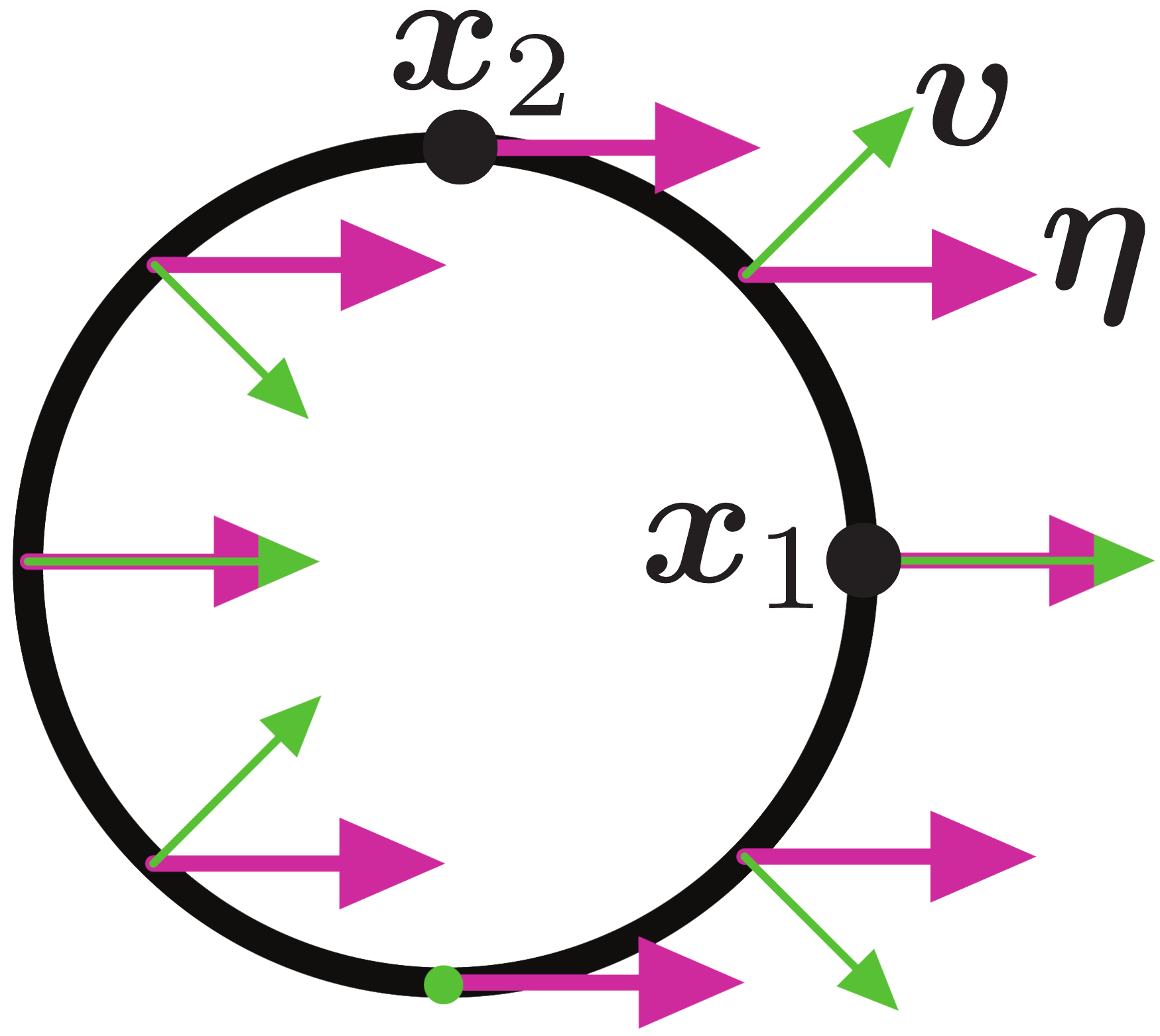}
  \captionsetup{justification=raggedright}
  \caption{ Non-rigid consistent field $\vv$.}\label{fig:circle}
\end{wrapfigure}
is a particular solution to the consistency equation, see \cref{fig:circle} (in green).
Now consider the two points on the sphere $\vx_1 = \frac{\veta}{\norm{\veta}}$ and $\vx_2 = \frac{\veta^\perp}{\norm{\veta}}$, where $\veta^\perp$ is an orthogonal vector to $\veta$, so that $\vw(\vx_1) = \veta$ while $\vw(\vx_2)=0$.
Thus, while the surface is moving rigidly with uniform velocity $\veta$, we have found an ``alternative'' velocity field $\vw$ that is compatible with the change in the implicit function, but that causes the distance between the surface points $\vx_1$ a $\vx_2$ to change.
We can recover the ``true'' velocity field by setting the auxiliary vector $\vu(\vx_i) = \veta$, which results in $\vv(\vx_i) = \vw(\vx_i) + \mP \vu(\vx_i) = \veta$ for $i=1,2$.

\paragraph{Killing energy.}

The Killing energy~\cite{solomon11as-killing-as-possible,tao2016near,slavcheva2017killingfusion} is applied to vector fields and measures their deviation from vector fields induced from rigid motions, making it an obvious choice of regularizer in our case.
The Killing energy of a vector field $\vv$ is defined as
\begin{equation}\label{e:rho_arap}
\rho(\vx;\vv) = \norm{\nabla_\vx \vv(\vx) + \nabla_\vx\vv(\vx)^T }^2_F.
\end{equation}

\paragraph{Rigid deformation prior.}

Next, we look at the problem of predicting the auxiliary vector field $\vu(\vx)$.
For this, we may use a sophisticated regressor such as a second neural network, but this is prone to overfitting.
From the example above, $\vu$ should approximate the velocity field of the object, which is locally rigid.
If the motion was globally rigid (\ie, $\vx(t) = R_t \vx(0) + T_t$ for time-varying $(R_t,T_t)\in SE(3)$), then it could be explained by an affine choice of velocity field:
% has two issues:
% Taking a complex $\vu$ (\eg, a second neural network) leads to two issues: 
% a challenging inner optimization in the loss in~\cref{e:loss_emp}, and high flexibility in reducing the deformation loss resulting in less efficient deformation prior \yl{XXX do we demonstrate this is results/supp? if so worth mentioning here}.
% A useful choice is to use affine vector fields $\vu$:
\begin{equation}\label{e:linear_u}
\vu(\vx)=\mA\vx+\vb,
\end{equation}
where $(\mA,\vb)\in\Real^{3\times 4}$ (this can be seen by differentiating $\vx(t)$ with respect to time).
Now plugging~\cref{e:linear_u} in~\cref{e:v} and taking the derivative w.r.t.~$\vx$ gives:
$$
\nabla_\vx \vv = \nabla_\vx \vw + (D_\vx\mP) (\mA\vx+\vb) + \mP \mA,$$
where, for a vector $\vq=(q_1,q_2,q_3)^T$, we denote $\brac{(D_\vx \mP)\vq}_i= \sum_{j=1}^3 \nabla_\vx P_{ij}(\vx) q_j$. Using this in~\cref{e:rho_arap,e:loss_emp} leads to a least-squares (LS) problem in $\mA,\vb$.
Minimization w.r.t.~$\mA,\vb$ is done by solving the normal equations.
Let $\mA_\star,\vb_\star$ denote the minimizer of this LS problem, so that $\vu_\star(\vx) = \mA_\star \vx +\vb_\star$. Plugging $\vu_\star$ in~\cref{e:loss_emp} yields
\begin{equation}\label{e:loss_emp_danskin}
    \loss_{\text{d}}(\theta) = \frac{1}{n}\sum_{i=1}^n \rho(\vx_i; \vv_{\vz_i,\veta_i,\vu_\star}).
\end{equation}
Clearly, the two losses in equations \cref{e:loss_emp_danskin} and \cref{e:loss_emp} share the same value at $\theta$.
Furthermore, Danskin's Theorem (see supplamentary) asserts that they also have the same gradient at $\theta$. 

\paragraph{Multiple rigid deformation priors.}

A single affine $\vu$ can explain a global rigid motion of the implicit shape.
Combining a collection of such affine fields can be used to introduce a piecewise rigid motion prior.
In order to do so, we introduce probability vector, also modeled with a neural network (we will continue to denote the collection of all learnable parameters by $\theta$), $\vp=(p_1,\ldots,p_k):\Real^3\too [0,1]^k$, where $k$ is a hyper-parameter and $\sum_{j=1}^k p_j(\vx)=1$ for all $\vx$.
We then introduce $k$ affine fields $\vu_j(\vx) = \mA_j\vx + \vb_j$, $j\in[k]$, and define our deformation loss as:
\begin{equation}\label{e:loss_emp_multi}
    \loss_{\text{d}}(\theta) = \min_{\vu_1\ldots\vu_k} \frac{1}{n}\sum_{j=1}^k\sum_{i=1}^n p_j(\vx_i) \rho(\vx_i; \vv_{\vz_i,\veta_i,\vu_j})
\end{equation}

% \yl{andrea new loss: written draftly for now}
% \begin{align*}
%     &\loss_{\text{d}}(\theta) = \\ &\min_{\vu_1\ldots\vu_k} \frac{1}{n}\sum_{i,j} p_j(\vx_i) \|\nabla_\vx f(\vx_i,\vz_i)\brac{(\mA-\mA^T)\vx_i + \vb} \\ & + \frac{\partial f}{\partial \vz}(\vx_i,\vz_i)\veta \|^2 
% \end{align*}
 
%\yl{end new loss}

This loss divides the surface $\gS(\vx)$ to (soft) pieces according to the probability vector $\vp$, and encourages each piece to be deformed with a rigid motion.
Minimization w.r.t.~$\vu_1,\ldots,\vu_k$ can be done independently for each $j\in[k]$, as in the case of a single rigid deformation prior.
The only difference is the incorporation of weights $p_j(\vx_i)$, leading to a weighted LS problem.
Using Danskin's theorem again will guarantee that plugging the minimizing fields $\mA_j^\star, \vb^\star_{j}$ in the loss in~\cref{e:loss_emp_multi} provides correct value and gradient at current parameter $\theta$.

\section{Implementation details}\label{s:implementation_details}

In addition to the deformation prior $\loss_\text{d}$, which is the main contribution of this work and described in the previous section, we use standard reconstruction losses for the shape space requirement (i), \ie, $\gS(\vz^{(i)})\approx\gX^{(i)}$, $i\in [m]$.
In this section we provide the remaining details on our loss and architecture. 

We use two networks in our system, the implicit representation $f:\Real^{3}\times\Real^D\too \Real$, and the probability network $\vp=(p_1,\ldots,p_k):\Real^{3}\times\Real^D\too \Real^k$. We use $k=20$ unless otherwise stated, an ablation test on $k$ is provided in the supplementary. 
Both networks are Multilayer Perceptrons (MLPs) (specific architectures are in the supplementary).  
$\vp(\vx,\vz)$ represents the parts' probabilities of a point $\vx$ in the shape $\gS(\vz)$.
We use the Auto-Decoder (AD) paradigm~\cite{Park_2019_CVPR}, where the latent codes $\gZ=\set{\vz^{(i)}}_{i\in [m]}$ are learnable parameters.
We incorporate the AD regularizer encouraging the latent codes to have a scaled Gaussian distribution (see~\cite{Park_2019_CVPR} for more details), 
\begin{equation}\label{e:loss_ae}
    \loss_{\text{ad}}(\theta) = \frac{1}{m}\sum_{i=1}^m \|\vz^{(i)}\|^2,
\end{equation}
where $\norm{\vz}=\sqrt{\vz^T\vz}$ is the standard euclidean norm, and we henceforth denote by $\theta$ the collection of all learnable parameters of the system, including the parameters of $f$, $\vp$, and $\gZ$.

\paragraph{Deformation loss.}

We use \cref{e:loss_emp_multi} as our deformation loss, where the samples $\vz_i,\veta_i,\vx_i$ used for each iteration are constructed as follows.
For each $i\in[n]$ in a batch of size $n$, we random a pair of latent codes  $\vz^{(i_1)},\vz^{(i_2)}$ from $\gZ$, and a scalar $t$ uniformly in $[0,1]$.
Next, we define $\vz_i$ to be the time $t$ interpolation of $\vz^{(i_1)}$ and $\vz^{(i_2)}$.
We have tested two options:
\emph{linear}, where $\vz_i=(1-t)\vz^{(i_{1})} + t\vz^{(i_{2})}$, and \emph{spiral}, where
$$
\vz_i\hspace{-2pt} = \hspace{-2pt}\brac{(1-t)\|\vz^{(i_{1})}\| + t\|\vz^{(i_{2})}\|}\text{slerp}\parr{\frac{\vz^{(i_{1})}}{\|\vz^{(i_{1})}\|},\frac{\vz^{(i_{2})}}{\|\vz^{(i_{2})}\|}}
$$
and $\text{slerp}$ is spherical linear interpolation.
We found the spiral interpolation favorable since it does not decrease the interpolated latent codes norm; we compare between the two options in the experimental section.
For each such $\vz_i$, we set $\veta_i$ to be the normalized speed $\frac{d \vz_i}{d t}$ of the interpolation, and $\vx_i$ a random sample from $\gS(\vz_i)$ computed by projecting a random point in space onto $\gS(\vz_i)$ as detailed in the supplementary. We also add the Eikonal loss \cite{gropp2020implicit}, denoted $\loss_\text{e}$, to regularize the level-sets of implicit surfaces at intermediate latent codes $\vz_i$, see supplementary for details. 

% \begin{equation}\label{e:eikonal}
%     \loss_{\text{e}}(\theta) = \frac{1}{n}\sum_{i=1}^n \parr{\norm{\nabla_\vx f(\vy_i,\vz_i)}-1}^2,
% \end{equation}
% where $\vy_i\in\Real^3$ are sampled uniformly in a bounding box of all the input shapes.

\begin{figure*}[t]
\centering
    % this is a smaller version if we need the space later....
    %\hspace{23pt}
     \begin{tabular}{@{\hskip0pt}c@{\hskip10pt}|@{\hskip10pt}c@{\hskip0pt}} 
                
                \includegraphics[width=0.45\textwidth]{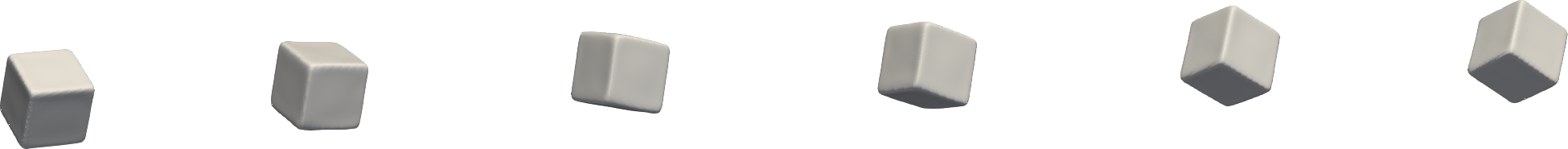} & \includegraphics[width=0.45\textwidth]{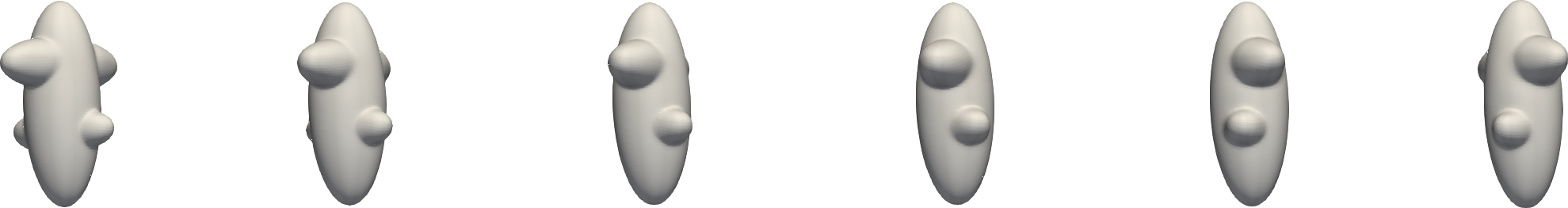}\\
                \includegraphics[width=0.45\textwidth]{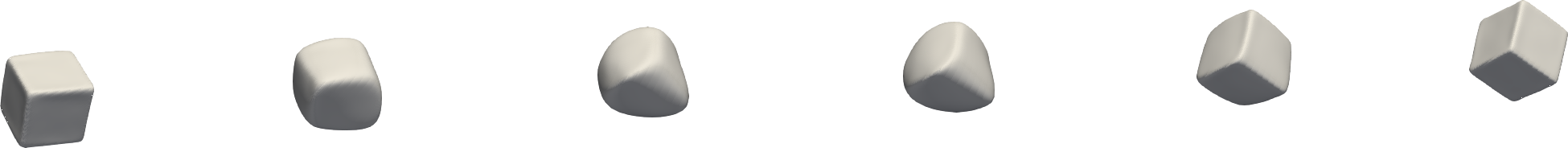} & \includegraphics[width=0.45\textwidth]{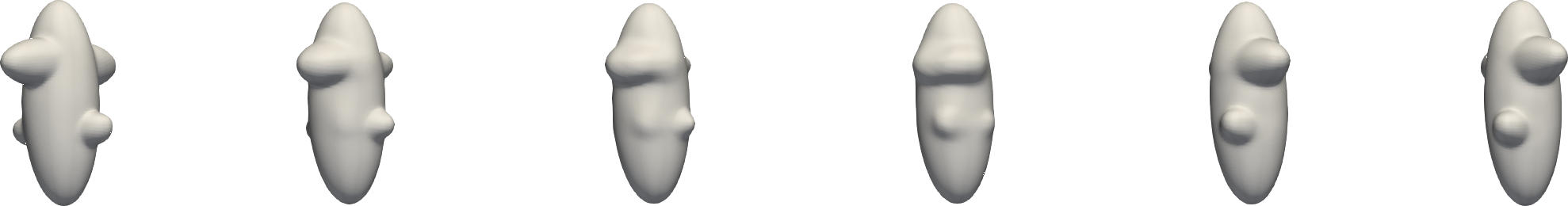}  
                %                       \\ %\hline
                %   \\                %\hline
            \end{tabular}
     
    % \begin{tabular}{@{\hskip0pt}c@{\hskip10pt}} 
    %             \includegraphics[width=0.9\textwidth]{figures/evaluation/ours_cube.png} \\ 
    %             \includegraphics[width=0.9\textwidth]{figures/evaluation/baseline_cube.png} \\
    %             \includegraphics[width=0.9\textwidth]{figures/evaluation/ours_ellipse.png}\\
    %             \includegraphics[width=0.9\textwidth]{figures/evaluation/baseline_ellipse.png}  
    %             %                       \\ %\hline
    %             %   \\                %\hline
    %         \end{tabular}
      
             \caption{Toy dataset: incorporating the deformation prior $\loss_\text{d}$ (top row) allows reconstructing natural shape interpolation, in contrast to baseline (bottom row).
             Left and right in each sequence are the reconstructed input shapes.
             }\vspace{-15pt}\label{fig:toy}
\end{figure*}

\paragraph{Reconstruction loss.}
In order to approximate the input shapes $\gX^{(i)}$ at the latent codes $\vz^{(i)}$, we use the SALD reconstruction loss \cite{atzmon2020sald}, denoted $\loss_\text{r}$, that handles raw data as input, and only requires the \emph{unsigned} distance to the input geometry, $d(\vx,\gX) = \min_{\vy\in\gX} \norm{\vx-\vy}$. See supplementary for details on this loss. 
% For a batch of size $b$ of samples $\vq_i, \vz^{(i)}$, $i\in[b]$, the loss is defined by
% \begin{align*}
%     \loss_{\text{r}}(\theta) &= \frac{1}{b} \sum_{i=1}^b \Big [ \tau\big(f(\vq_i,\vz^{(i)}),d(\vq_i,\gX^{(i)})\big) \\  &+ \lambda \tau\big(\nabla_\vx f(\vq_i, \vz^{(i)}), \nabla_\vx d(\vq_i,\gX^{(i)})\big)\Big ],
% \end{align*}
% where $\tau$ is the sign agnostic function, \ie, $\tau(a,b) = \min\set{|a-b|,|a+b|}$ for scalars, and $\tau(\va,\vb)=\min\set{\norm{\va-\vb},\norm{\va+\vb}}$ for vectors.
% The latent $\vz^{(i)}$ is a random sample from $\gZ$, and the point $\vq_i$ is sampled by adding Gaussian noise to a uniform sample from the input geometry $\gX^{(i)}$ (see supplementary and \cite{atzmon2020sald} for more details).
% Lastly, $\lambda=0.1$.
%
%
Our complete loss function is 
\begin{equation}\label{e:loss_complete}
    \loss = \loss_{\text{r}} + \lambda_\text{d}\loss_\text{d} +  \lambda_\text{e}\loss_{\text{e}} + \lambda_\text{ad}\loss_\text{ad},
\end{equation}
where the losses are all functions of the parameters $\theta$, and we use $\lambda_\text{e}=0.1$, and $\lambda_\text{ad}=0.001$. For $\lambda_\text{d}$ we use scheduling, details in the supplementary.

\section{Experiments}

\paragraph{Baselines.}

For baselines we picked two popular implicit shape space learning methods that have been shown to produce state of the art results on the same or similar datasets:
(i) Auto-Decoder (AD)~\cite{Park_2019_CVPR}; and 
(ii) Variational Auto-Encoder (VAE)~\cite{mescheder2019occupancy}.
In both cases, we trained using the SALD reconstruction loss \cite{atzmon2020sald}.
For the encoder in the VAE, we use PointNet~\cite{qi2017pointnet}, similar to previous works that utilizing VAE on this task~\cite{mescheder2019occupancy,atzmon2019sal,atzmon2020sald}. The full details of the network architecture can be found in supplementary.

\subsection{Evaluation using toy data}\label{s:evaluation}

In the first experiment, we test the ability of our model to learn from a synthetic shape space with known and well understood deformations between shapes.
To this end, we generated two ``toy'' datasets:
(i) 12 cubes, randomly rotated and translated in 3D space; 
and (ii) 12 ellipsoids (``main body'') with upper and bottom perturbing part (``arms'' and ``legs''), where the main body ellipsoid is randomly translated in the $x$-axis direction and the arms and legs ellipsoids are deformed by two different random rotations around the $z$-axis.
For each pair of shapes in this dataset there exists a two-piece rigid deformation field, deforming one shape to the other.
See the supplementary for a video showing few samples of shapes from the two datasets. 

On each of these datasets we trained:
(i) Vanilla auto-decoder, \ie, without our deformation prior loss, $\lambda_\text{d}=0$; and
(ii) using the same auto-decoder architecture and loss as in (i), but now with active deformation loss, \ie, $\lambda_d=0.001$.
We set $k$ to 1 for the cubes and 2 for the ellipses to match the ground truth number of parts.

The bottom and upper row in \cref{fig:toy} show learned implicit shapes for $\lambda_d=0$ and $\lambda_d=0.001$, respectively.
In each sequence, the leftmost and rightmost shapes correspond to a pair of learned latent codes $\vz^{(i_1)},\vz^{(i_2)}$ , $i_1,i_2\in[12]$.
Note that both methods learn to reconstruct the input data well.
The middle shapes in this experiment were generated using linear interpolation of latent codes, $(1-t)\vz^{(i_1)} + t\vz^{(i_{2})}$, $t\in \left[0.15,0.35,0.65,0.85 \right]$.
Note, in particular, that the shapes generated with $\lambda_d=0$ fail to keep important geometrical features and revert to smooth but ``non-geometrical'' interpolation.
This can be seen from the smoothing of the cubes and the blending of the arms and legs in the ellipsoid.
In contrast, enabling the deformation loss leads to geometrically-plausible interpolations, close to the ground-truth ones.

\subsection{Shape space learning}\label{s:shape_space}

Next, we test our deformation prior on the task of learning a shape space from ``real-life'' raw scans.
We considered the D-Faust raw scans dataset \cite{bogo17dynamic}, consisting of 41k scans from 10 humans: 5 male and 5 female.
The data for each individual is categorized into sequences of actions.
Together with the raw data, mesh registrations are also in the dataset.
We use the registrations for evaluation, but not for training.

\begin{table}[t]
    \centering
    \scriptsize
    \setlength\tabcolsep{2pt} % default value: 6pt
    \begin{tabular}{c}
        \begin{adjustbox}{max width=\textwidth}
            \aboverulesep=0ex
            \belowrulesep=0ex
            \renewcommand{\arraystretch}{1.1}
            \begin{tabular}[t]{c|c|c|c | c | c|}
            \multicolumn{2}{c}{} & 
            \multicolumn{2}{|c}{Registrations} & 
            \multicolumn{2}{|c}{Scans} \\
                \cmidrule{2-6}
                & Method & Chamfer & Wasserstein & Chamfer & Wasserstein \\
                \midrule
                \multirow{ 3}{*}{Unique action} &
                Ours  &  \textbf{0.112} & \textbf{2.994} & \textbf{0.098} &  \textbf{2.971} \\
                & AD  & 0.231 & 3.527 & 0.195 & 3.354  \\
                & VAE & 0.237 & 3.361 & 0.146 & 3.226 \\
                 \midrule
                 \multirow{ 3}{*}{Punching} &
                 Ours & \textbf{0.126} & \textbf{3.164} & \textbf{0.103} & \textbf{3.069} \\
                & AD & 0.995 & 3.757 & 0.198 & 3.671  \\
                & VAE & 0.656 & 4.558 & 0.613  & 4.351 \\
                \midrule
                \multirow{ 3}{*}{One leg jump} &
                Ours  &  \textbf{0.278} & \textbf{3.490} &\textbf{1.313} &  \textbf{5.601} \\
                
                & AD  & 0.385 & 3.961 & 1.414 & 5.908  \\
                & VAE & 0.957 & 5.077 & 1.992 &  7.064 \\
                 \midrule
                 \multirow{ 3}{*}{Light hopping} &
                 Ours & 0.053 & \textbf{2.801} & 0.045 & 2.871  \\
                & AD & \textbf{0.050} & 2.818 & \textbf{0.041} & \textbf{2.857}  \\
                & VAE & 0.086 & 2.906 & 0.081   & 2.903 \\
                 \midrule
                 \multirow{ 3}{*}{Shake hips} &
                 Ours  & \textbf{0.097} & \textbf{3.016} & \textbf{0.211} & \textbf{3.358}  \\
                & AD & 0.128 & 3.228 & 0.244 & 3.532  \\
                & VAE & 0.217 &  3.258 & 0.343  & 3.469 \\
                \cmidrule{2-6}
                %\bottomrule
        \end{tabular} 
        \end{adjustbox}
      
    \end{tabular}
    %\vspace{3pt}
    \caption{Reconstruction of test set of all humans (splits 1 and 2) in D-Faust.  We log Chamfer and Wasserstein distances of the reconstructed surfaces to the raw scans and ground-truth registrations; Chamfer reported $*10^3$. \vspace{-10pt}}
    \label{tab:all_humans_splits}
\end{table}

\begin{table}[t]
    \centering
    \scriptsize
    \setlength\tabcolsep{2pt} % default value: 6pt
    \begin{tabular}{c}
        \begin{adjustbox}{max width=\textwidth}
            \aboverulesep=0ex
            \belowrulesep=0ex
            \renewcommand{\arraystretch}{1.1}
            \begin{tabular}[t]{c|c|c|c | c | c|}
            \multicolumn{2}{c}{} & 
            \multicolumn{2}{|c}{Registrations} & 
            \multicolumn{2}{|c}{Scans} \\
                \cmidrule{2-6}
                & Method & Chamfer & Wasserstein & Chamfer & Wasserstein \\
                \midrule
                \multirow{ 3}{*}{Male random} &
                Ours (spiral) &  0.073 & \textbf{3.329} & 0.048 &  \textbf{3.233} \\
                & Ours (linear)  & 0.080 & 3.360 & 0.051 & 3.328  \\
                & AD  & \textbf{0.071} & 3.361 & \textbf{0.045} & 3.292  \\
                & VAE & 0.128 & 3.503 & 0.164  & 3.457 \\
                 \midrule
                 \multirow{ 3}{*}{Female random} &
                 Ours (spiral) & 0.042 & 2.490 & 0.042 & 2.530 \\
                & Ours (linear) & \textbf{0.036} & \textbf{2.450} & \textbf{0.036} & \textbf{2.470}  \\
                & AD & 0.056 & 2.600 & 0.049 & 2.640  \\
                & VAE & 0.163 & 2.841 & 0.152 & 2.842 \\
                \midrule
                \multirow{ 3}{*}{Male action} &
                Ours (spiral) &  1.017 & 6.410 &0.822 &  5.360 \\
                & Ours (linear)  & \textbf{0.762} & \textbf{5.499} & \textbf{0.649} &  \textbf{4.616}  \\
                & AD  & 1.201 & 6.657 & 0.979 & 5.613  \\
                & VAE & 2.443 & 9.171 & 2.270  & 8.182 \\
                 \midrule
                 \multirow{ 3}{*}{Female action} &
                 Ours (spiral) & \textbf{0.048} & \textbf{2.518} & \textbf{0.042} & \textbf{2.456} \\
                & Ours (linear) & 0.052 & 2.633 & 0.045 & 2.620  \\
                & AD & 0.068 & 2.806 & 0.059 & 2.709  \\
                & VAE & 8.509 & 17.024 & 8.547  & 16.951 \\
                \cmidrule{2-6}
                %\bottomrule
        \end{tabular} 
        \end{adjustbox}
    \end{tabular}
    %\vspace{3pt}
    \caption{Reconstruction of test set of individuals (splits 3 and 4) in D-Faust.  We log the Chamfer and Wasserstein distances of the reconstructed surfaces to the raw scans and ground-truth registrations; Chamfer reported $* 10^3$. } % \vspace{-15pt}
    \label{tab:individual_splits}
\end{table}

\paragraph{Shape space reconstruction.}

To test the ability of our learned shape space to generalize and represent unseen geometries from sparse training examples, we have created the following data splits from the D-Faust dataset:
% (1) \emph{Unique action}: here we removed a unique action from each of the 10 humans as test, and took 1 out of every remaining 5 scans for training;
% (2) \emph{Punching}/\emph{One leg jump}/\emph{Light hopping}/\emph{Shake hips}: here we removed the same action from all 10 humans as test, and took 1 out every remaining 5 for training;
% (3) \emph{Male action}/\emph{Female action}: here we took one action from each as test, and out of every remaining 10 sampled 1 for training; and
% (4) \emph{Male random}/\emph{Female random}: here we used the same training set as (3) but took as test random scans from the left out 9 out of 10 scans.
% In general, since all these data splits are still very large, we randomly subsampled 5\%-10\% of the scans for experimentation. \ma{i think we should remove the last sentence}
% DN: rephrased below, feel free to revert to the original version above
(1) \emph{Unique action}: removes a unique action from each of the 10 human identities as a test set, and trains on every 5-th scan from the remaining time-ordered scans;
(2) \emph{Punching}/\emph{One leg jump}/\emph{Light hopping}/\emph{Shake hips}: removes the same action from all 10 humans as a test set, and trains on every 5th scan of the time-ordered remainder.
(3) \emph{Male action}/\emph{Female action}: takes one action from each human as a test set, and out of a tuple $\mathcal{F}$ of all remaining ordered frames takes every 10-th for training.
(4) \emph{Male random}/\emph{Female random}: uses the same training set as (3), but from every 10 consecutive frames in $\mathcal{F}$, randomly samples from frames 1-9 for testing.
In general, since all these data splits are still very large, we randomly subsampled 5\%-10\% of the scans for experimentation. 

% so 4) means, 1) From each human we remove an action to create a set of frames F; 2) From F we sample every 10th frame to generate a training set F_train; 3) From the F \ F_train we sample random scans for F_test?

%the final test sets used. 

\begin{figure}[t]
\centering
     \includegraphics[width=0.4\textwidth]{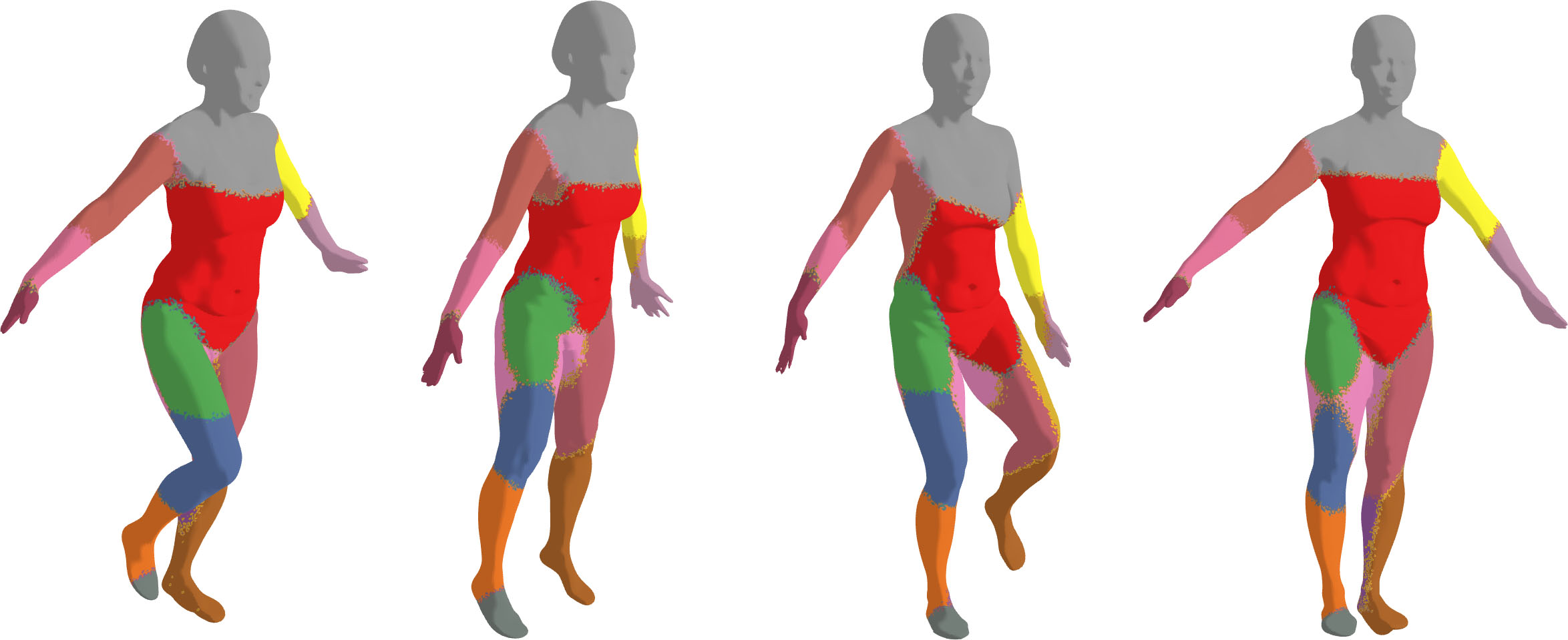}
    \caption{The learned probabilities $\vp$ for several reconstructions from the female split (splits 3 and 4).}
    \label{fig:prob}
\end{figure}

To test how well the learned shape space represents unseen test cases, we follow~\cite{Park_2019_CVPR} and, for each test scan  $\gX$, we minimize the reconstruction loss with respect to the latent code $\vz\in\Real^D$ with the weights of the neural network $f$ fixed. 
% \Cref{tab:all_humans_splits} reports the results of splits 1 and 2 (all humans), and \cref{tab:individual_splits} reports the result of splits 3 and 4 (individuals).
\Cref{tab:all_humans_splits} and \cref{tab:individual_splits} report the results on splits 1 and 2 (all humans), and on splits 3 and 4 (individuals) respectively.
In the latter, we also provide a comparison between linear and spiral interpolations.
Figure \ref{fig:test_recon} shows some typical results from splits 1 and 2; we show failure cases in the supplementary. In order to measure reconstruction accuracy we use the Chamfer and Wasserstein distances, as customary (details in the supplementary).
As can be seen by inspecting the quantitative and qualitative results, our deformation prior indeed improves prediction ability to unseen scans representing a missing action or pose. For example, when comparing the red boxes in Figure \ref{fig:test_recon} to the blue boxes (unseen ground truth) our method is able to reconstruct the most similar pose, outperforming the baselines.  
Figure \ref{fig:prob} depicts the probability vectors (a different color for each of the $k=20$ pieces) for reconstructed examples of the Female (splits 3 and 4).
Interestingly, the vectors correspond to the natural kinematic structure of the body.

\begin{figure}[t]
    % \newcommand{\testrconmethodbox}[1]{\parbox{0.2\linewidth}{\centering{#1}}}
    % \parbox{0.3\linewidth}{\centering Unseen scans}%
    % \testrconmethodbox{Raw scan}%
    % \testrconmethodbox{Ours}%
    % \testrconmethodbox{AD}%
    % \testrconmethodbox{VAE}\\
     \includegraphics[width=0.48\textwidth]{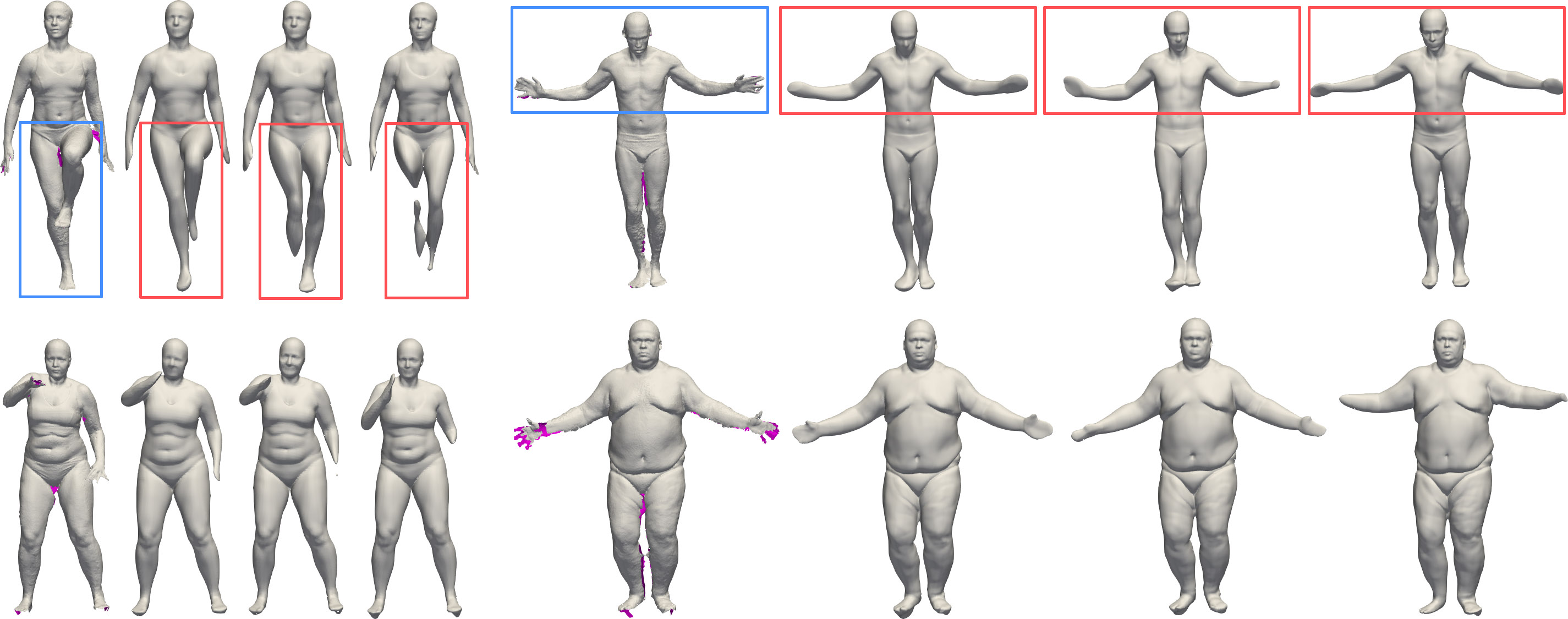}
    \caption{Shape space reconstruction of unseen poses (taken from splits 1,2 and 4). From left to right (in each sequence of 4): unseen raw scan, ours, AD, VAE.}
    \label{fig:test_recon}
\end{figure}

% baselines: autodecoder [deepSDF], VAE [SAL,SALD].

% dataset: 6 sparse splits of all D-Faust (remove unique action, remove same same action), 2 very sparse splits on 2 humans (random frames, action), all cat (random).

% \begin{enumerate}
% \item toys: cubes, ellipses (4x4 grids)
% \item ablation: on $k$, and visualization of probabilities. 
% \item shape space evaluation: table, figure examples (one column). 
% \item interpolation: small table (with linear and slerp) + figures (wide column) of sequence with baseline. 

% \end{enumerate}

\begin{figure*}[t]
% \newcommand{\methodheader}[1]{\parbox{3.5cm}{\centering#1}}
% \rotatebox{-90}{%
% \methodheader{Ours}%
% \methodheader{AD}%
% \methodheader{VAE}%

% \methodheader{Ours}%
% \methodheader{AD}%
% \methodheader{VAE}%
% }
    \hspace{23pt}
\centering
            \begin{tabular}{@{\hskip0pt}c@{\hskip5pt}|@{\hskip5pt}c@{\hskip0pt}} 
\includegraphics[width=0.515\textwidth]{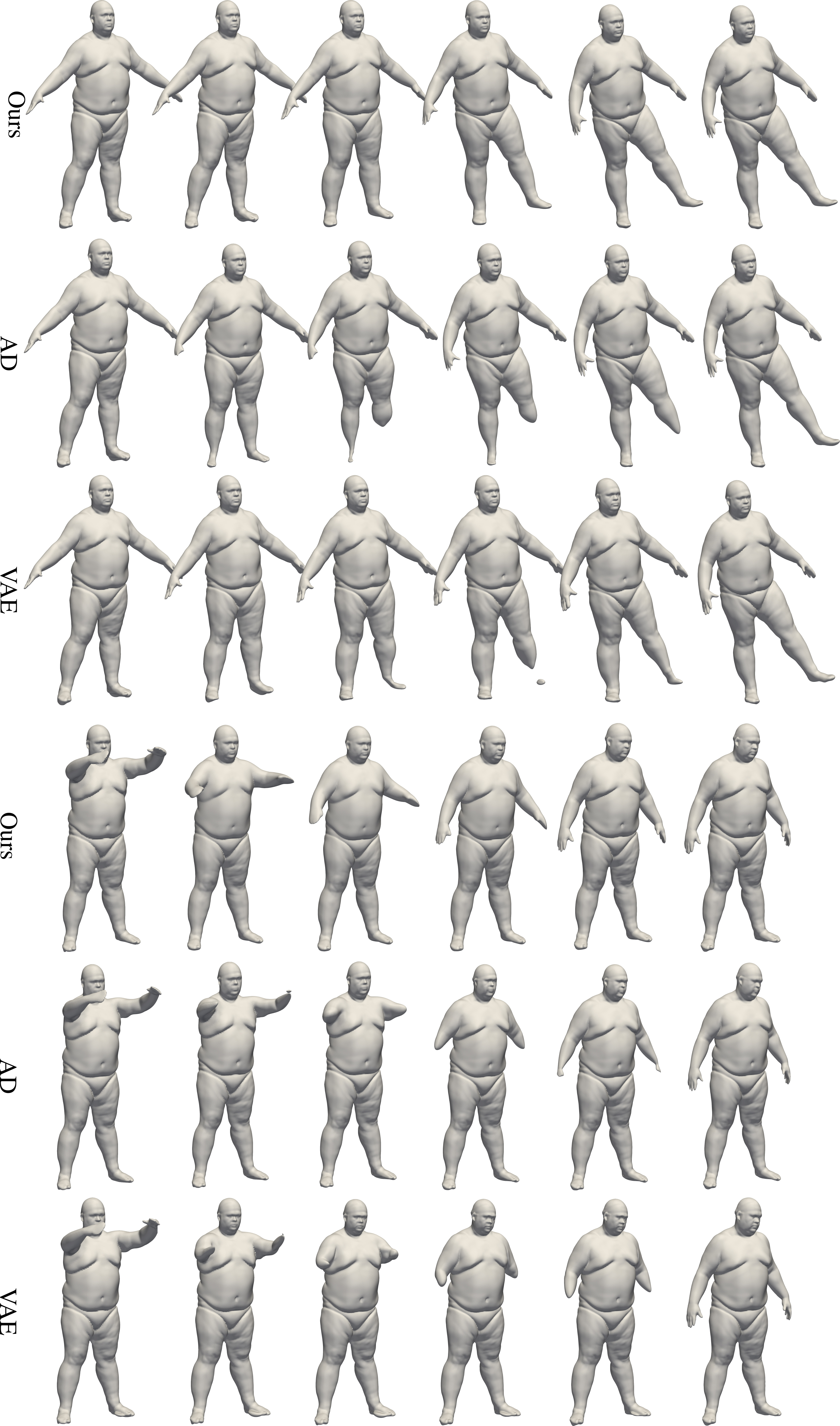} & \includegraphics[width=0.45\textwidth]{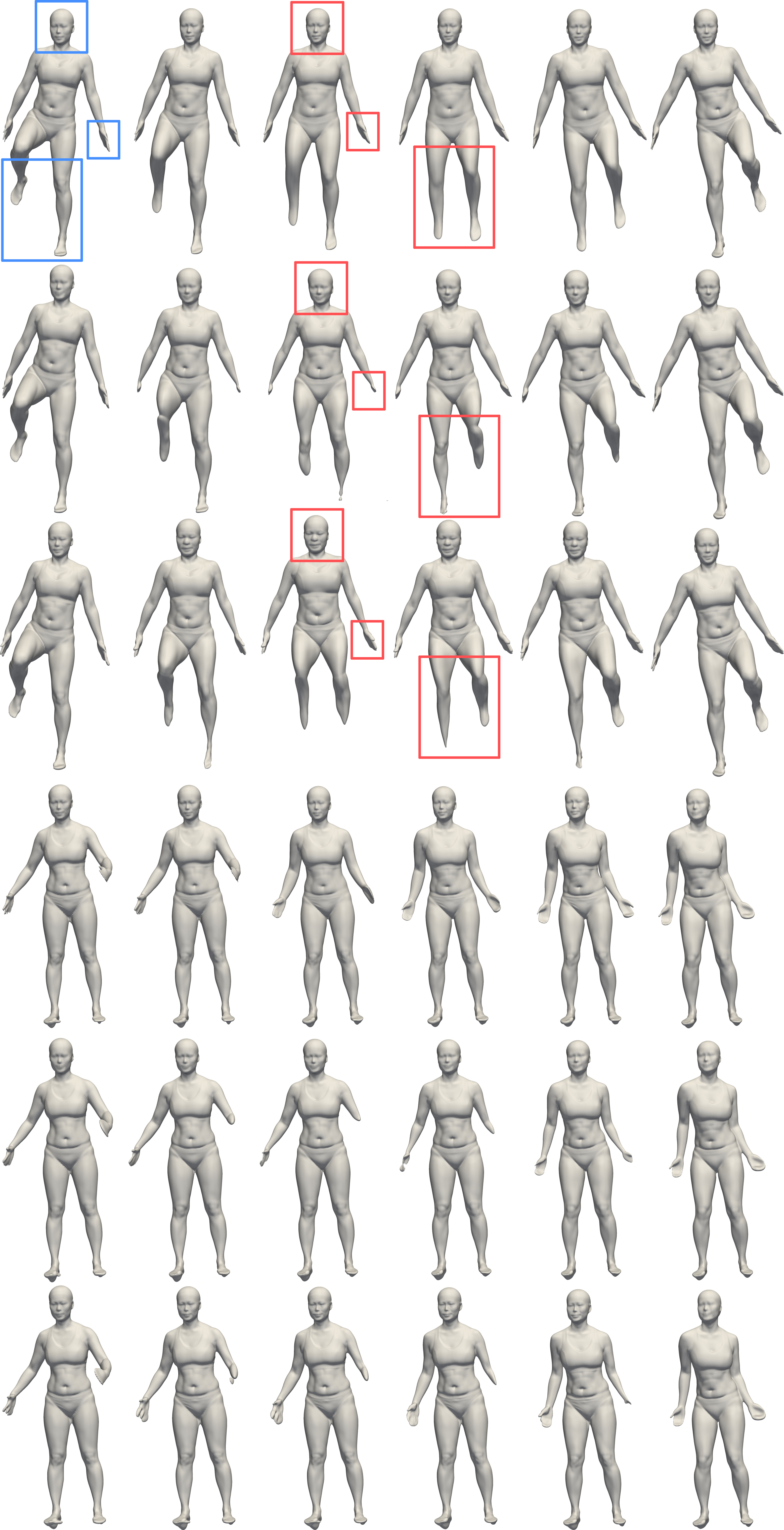}
            \end{tabular}
             \caption{Interpolation of latent codes. Each block of three rows shows (top to bottom): Our result, AD, and VAE. Note the natural shape and motion of the intermediate shapes when incorporating our deformation prior. }%\vspace{-15pt}
            \label{fig:interpolation}
\end{figure*}

% \begin{figure*}[th]
%     \hspace{23pt}
% \centering
%             \newcommand{\methodheader}[1]{\parbox{2.53cm}{\scriptsize\centering#1}}%
%             \begin{tabular}{c@{\hskip0pt}c@{\hskip10pt}|@{\hskip10pt}c@{\hskip10pt}}%
%             \rotatebox[origin=r]{-90}{%
%             \methodheader{Ours}%
%             \methodheader{AD}%
%             \methodheader{VAE}%
%             \methodheader{Ours}%
%             \methodheader{AD}%
%             \methodheader{VAE}%
%             }&%
%             \includegraphics[width=0.49\textwidth]{figures/one_human/50002.jpg} & \includegraphics[width=0.44\textwidth]{figures/one_human/50020.jpg}%
%             \end{tabular}
%              \caption{Interpolation of latent codes. Each block of three rows shows (top to bottom): Our result, AD, and VAE. Note the natural shape and motion of the intermediate shapes when incorporating our deformation prior. }%\vspace{-15pt}
%             \label{fig:interpolation}
% \end{figure*}

\begin{table}[t]
    \centering
    \scriptsize
    \setlength\tabcolsep{2pt} % default value: 6pt
    \begin{tabular}{c}
        \begin{adjustbox}{max width=\textwidth}
            \aboverulesep=0ex
            \belowrulesep=0ex
            \renewcommand{\arraystretch}{1.1}
            \begin{tabular}[t]{c|c|c|c | c | c| c}
                & Method & HKS($t_{\text{min}}$) & HKS($t_{\text{mean}}$) & HKS($t_{\text{max}}$) & area & geodesic \\
                \midrule
                \multirow{ 3}{*}{Male} &
                Ours (slerp) & 955.422 & \textbf{4.804} & \textbf{4.775} &  \textbf{120.632} & \textbf{0.717} \\
                & Ours (linear)  & \textbf{775.381} & 4.946 & 4.914 & 179.856 & 0.977  \\
                & AD  & 1004.734 & 6.150 & 6.104 & 189.676 & 1.119  \\
                 \midrule
                 \multirow{ 3}{*}{Female} &
                 Ours (slerp) & \textbf{191.823} & \textbf{6.511} & \textbf{6.433} & \textbf{106.807} & \textbf{0.836} \\
                & Ours (linear) & 239.745 & 7.664 & 7.584 & 143.439 & 1.073  \\
                & AD & 468.887 & 14.830 & 14.669 & 151.859 & 1.203  \\
                \cmidrule{2-7}
                %\bottomrule
        \end{tabular} 
        \end{adjustbox}
    \end{tabular}
    %\vspace{3pt}
    \caption{Evaluation of interpolated surfaces. Wasserstein distance between histograms of different surface properties; HKS numbers are $* 10^6$; area and geodesic $*10^3$. } %\vspace{-15pt}
    \label{tab:eval_interpolation}
\end{table}

\subsection{Shape space interpolation}
We evaluated shape space interpolations on the individual splits (splits 3 and 4). The interpolations are using the spiral interpolation of learned latent codes $\gZ$. Figure \ref{fig:interpolation} presents qualitative results; note that our deformation prior improves intermediate poses and alleviates shortening of limbs that are common with the baselines.  For example, note that the highlighted face, hand, and legs in the red boxes are either distorted or deformed in the baselines compared to the interpolation end points (blue boxes), while our method maintains the geometric properties of these parts. For qualitative assessment of the interpolation results, we have compared histograms of different surface properties such as Heat Kernel Signature (HKS) at times $t_\text{min}$, $t_\text{mean}=0.5(t_\text{min}+t_\text{max}$), and $t_\text{max}$ as suggested in \cite{sun2009concise}, surface area, and pairwise geodesic distances \cite{osada2002shape}. In Table \ref{tab:eval_interpolation} we log the Wasserestein distances between histograms of interpolated surfaces and the corresponding histograms of the reconstructed shapes generated the interpolation (\ie, far left and right in each sequence in Figure \ref{fig:interpolation}). Note the advantage in using spiral interpolation method. 

\section{Conclusions}\label{s:conclusions}

In this paper we have analyzed the ability of deep neural implicit surfaces to interpolate between members of a set of complex 3D shapes, such as articulated human bodies. 
% To this end, we proposed a novel framework that regularizes the deformations in a shape space defined by neural implicit shape representations. 
To this end, we proposed a novel framework that regularizes the deformations within an implicit neural shape space.
The latter is done by first expressing the 3D deformation fields generated by an infinitesimal change of a latent shape code, followed by regularizing these  fields to be a collection of approximately rigid motions with an as-rigid-as-possible regularizer.
The experimental section has demonstrated that our framework significantly improves the generalization of the state of the art implicit shape representation networks especially when trained on a sparse and temporally non-smooth collection of 3D shapes, such as a heavily subsampled version of the raw D-Faust dataset.

{\small\bibliographystyle{ieee_fullname} \bibliography{egbib,vedaldi_general,vedaldi_specific}}

\begin{thebibliography}{10}\itemsep=-1pt

\bibitem{achlioptas2018learning}
Panos Achlioptas, Olga Diamanti, Ioannis Mitliagkas, and Leonidas Guibas.
\newblock Learning representations and generative models for 3d point clouds.
\newblock In {\em International conference on machine learning}, pages 40--49.
  PMLR, 2018.

\bibitem{atzmon2019controlling}
Matan Atzmon, Niv Haim, Lior Yariv, Ofer Israelov, Haggai Maron, and Yaron
  Lipman.
\newblock Controlling neural level sets.
\newblock In {\em Advances in Neural Information Processing Systems}, pages
  2032--2041, 2019.

\bibitem{atzmon2019sal}
Matan Atzmon and Yaron Lipman.
\newblock Sal: Sign agnostic learning of shapes from raw data.
\newblock In {\em IEEE/CVF Conference on Computer Vision and Pattern
  Recognition (CVPR)}, June 2020.

\bibitem{atzmon2020sald}
Matan Atzmon and Yaron Lipman.
\newblock Sald: Sign agnostic learning with derivatives.
\newblock {\em arXiv preprint arXiv:2006.05400}, 2020.

\bibitem{bacsar2008h}
Tamer Ba{\c{s}}ar and Pierre Bernhard.
\newblock {\em H-infinity optimal control and related minimax design problems:
  a dynamic game approach}.
\newblock Springer Science \& Business Media, 2008.

\bibitem{ben2010discrete}
Mirela Ben-Chen, Adrian Butscher, Justin Solomon, and Leonidas Guibas.
\newblock On discrete killing vector fields and patterns on surfaces.
\newblock In {\em Computer Graphics Forum}, volume~29, pages 1701--1711. Wiley
  Online Library, 2010.

\bibitem{dfaust:CVPR:2017}
Federica Bogo, Javier Romero, Gerard Pons-Moll, and Michael~J. Black.
\newblock Dynamic {FAUST}: {R}egistering human bodies in motion.
\newblock In {\em IEEE Conf. on Computer Vision and Pattern Recognition
  (CVPR)}, July 2017.

\bibitem{bogo17dynamic}
Federica Bogo, Javier Romero, Gerard Pons{-}Moll, and Michael~J. Black.
\newblock Dynamic {FAUST:} registering human bodies in motion.
\newblock In {\em Proc. {CVPR}}, 2017.

\bibitem{chen2019learning}
Zhiqin Chen and Hao Zhang.
\newblock Learning implicit fields for generative shape modeling.
\newblock In {\em Proceedings of the IEEE Conference on Computer Vision and
  Pattern Recognition}, pages 5939--5948, 2019.

\bibitem{chibane2020neural}
Julian Chibane, Gerard Pons-Moll, et~al.
\newblock Neural unsigned distance fields for implicit function learning.
\newblock {\em Advances in Neural Information Processing Systems}, 33, 2020.

\bibitem{choy20163d}
Christopher~B Choy, Danfei Xu, JunYoung Gwak, Kevin Chen, and Silvio Savarese.
\newblock 3d-r2n2: A unified approach for single and multi-view 3d object
  reconstruction.
\newblock In {\em European conference on computer vision}, pages 628--644.
  Springer, 2016.

\bibitem{cosmo2020limp}
Luca Cosmo, Antonio Norelli, Oshri Halimi, Ron Kimmel, and Emanuele
  Rodol{\`{a}}.
\newblock Limp: Learning latent shape representations with metric preservation
  priors.
\newblock In {\em Computer Vision - {ECCV} 2020 - 16th European Conference,
  Glasgow, UK, August 23-28, 2020, Proceedings, Part {III}}, volume 12348,
  pages 19--35. Springer, 2020.

\bibitem{eisenberger2020hamiltonian}
Marvin Eisenberger and Daniel Cremers.
\newblock Hamiltonian dynamics for real-world shape interpolation.
\newblock In {\em Computer Vision - {ECCV} 2020 - 16th European Conference,
  Glasgow, UK, August 23-28, 2020, Proceedings, Part {IV}}, volume 12349, pages
  179--196. Springer, 2020.

\bibitem{eisenberger2019divergence}
Marvin Eisenberger, Zorah L{\"a}hner, and Daniel Cremers.
\newblock Divergence-free shape correspondence by deformation.
\newblock In {\em Computer Graphics Forum}, volume~38, pages 1--12. Wiley
  Online Library, 2019.

\bibitem{genova2019deep}
Kyle Genova, Forrester Cole, Avneesh Sud, Aaron Sarna, and Thomas Funkhouser.
\newblock Local deep implicit functions for 3d shape.
\newblock In {\em Proceedings of the IEEE/CVF Conference on Computer Vision and
  Pattern Recognition}, pages 4857--4866, 2020.

\bibitem{genova2019learning}
Kyle Genova, Forrester Cole, Daniel Vlasic, Aaron Sarna, William~T Freeman, and
  Thomas Funkhouser.
\newblock Learning shape templates with structured implicit functions.
\newblock In {\em Proceedings of the IEEE International Conference on Computer
  Vision}, pages 7154--7164, 2019.

\bibitem{girdhar2016learning}
Rohit Girdhar, David~F Fouhey, Mikel Rodriguez, and Abhinav Gupta.
\newblock Learning a predictable and generative vector representation for
  objects.
\newblock In {\em European Conference on Computer Vision}, pages 484--499.
  Springer, 2016.

\bibitem{gropp2020implicit}
Amos Gropp, Lior Yariv, Niv Haim, Matan Atzmon, and Yaron Lipman.
\newblock Implicit geometric regularization for learning shapes.
\newblock In {\em Proceedings of Machine Learning and Systems 2020}, 2020.

\bibitem{groueix20183d}
Thibault Groueix, Matthew Fisher, Vladimir~G Kim, Bryan~C Russell, and Mathieu
  Aubry.
\newblock 3d-coded: 3d correspondences by deep deformation.
\newblock In {\em Proceedings of the European Conference on Computer Vision
  (ECCV)}, pages 230--246, 2018.

\bibitem{groueix2018papier}
Thibault Groueix, Matthew Fisher, Vladimir~G Kim, Bryan~C Russell, and Mathieu
  Aubry.
\newblock A papier-m{\^a}ch{\'e} approach to learning 3d surface generation.
\newblock In {\em Proceedings of the IEEE conference on computer vision and
  pattern recognition}, pages 216--224, 2018.

\bibitem{jiang2020shapeflow}
Chiyu Jiang, Jingwei Huang, Andrea Tagliasacchi, Leonidas Guibas, et~al.
\newblock Shapeflow: Learnable deformations among 3d shapes.
\newblock {\em arXiv preprint arXiv:2006.07982}, 2020.

\bibitem{kingma2014adam}
Diederik~P Kingma and Jimmy Ba.
\newblock Adam: A method for stochastic optimization.
\newblock {\em arXiv preprint arXiv:1412.6980}, 2014.

\bibitem{li2019supervised}
Lingxiao Li, Minhyuk Sung, Anastasia Dubrovina, Li Yi, and Leonidas~J Guibas.
\newblock Supervised fitting of geometric primitives to 3d point clouds.
\newblock In {\em Proceedings of the IEEE Conference on Computer Vision and
  Pattern Recognition}, pages 2652--2660, 2019.

\bibitem{lun20173d}
Zhaoliang Lun, Matheus Gadelha, Evangelos Kalogerakis, Subhransu Maji, and Rui
  Wang.
\newblock 3d shape reconstruction from sketches via multi-view convolutional
  networks.
\newblock In {\em 2017 International Conference on 3D Vision (3DV)}, pages
  67--77. IEEE, 2017.

\bibitem{mescheder2019occupancy}
Lars Mescheder, Michael Oechsle, Michael Niemeyer, Sebastian Nowozin, and
  Andreas Geiger.
\newblock Occupancy networks: Learning 3d reconstruction in function space.
\newblock In {\em Proceedings of the IEEE Conference on Computer Vision and
  Pattern Recognition}, pages 4460--4470, 2019.

\bibitem{mildenhall20nerf:}
Ben Mildenhall, Pratul~P. Srinivasan, Matthew Tancik, Jonathan~T. Barron, Ravi
  Ramamoorthi, and Ren Ng.
\newblock {NeRF}: Representing scenes as neural radiance fields for view
  synthesis.
\newblock In {\em Proc. {ECCV}}, 2020.

\bibitem{newcombe2011kinectfusion}
Richard~A Newcombe, Shahram Izadi, Otmar Hilliges, David Molyneaux, David Kim,
  Andrew~J Davison, Pushmeet Kohi, Jamie Shotton, Steve Hodges, and Andrew
  Fitzgibbon.
\newblock Kinectfusion: Real-time dense surface mapping and tracking.
\newblock In {\em 2011 10th IEEE International Symposium on Mixed and Augmented
  Reality}, pages 127--136. IEEE, 2011.

\bibitem{niemeyer19occupancy}
Michael Niemeyer, Lars~M. Mescheder, Michael Oechsle, and Andreas Geiger.
\newblock Occupancy flow: 4d reconstruction by learning particle dynamics.
\newblock In {\em Proc. {ICCV}}, 2019.

\bibitem{novotny2017learning}
David Novotny, Diane Larlus, and Andrea Vedaldi.
\newblock Learning 3d object categories by looking around them.
\newblock In {\em Proceedings of the IEEE International Conference on Computer
  Vision}, pages 5218--5227, 2017.

\bibitem{osada2002shape}
Robert Osada, Thomas Funkhouser, Bernard Chazelle, and David Dobkin.
\newblock Shape distributions.
\newblock {\em ACM Transactions on Graphics (TOG)}, 21(4):807--832, 2002.

\bibitem{Park_2019_CVPR}
Jeong~Joon Park, Peter Florence, Julian Straub, Richard Newcombe, and Steven
  Lovegrove.
\newblock Deepsdf: Learning continuous signed distance functions for shape
  representation.
\newblock In {\em The IEEE Conference on Computer Vision and Pattern
  Recognition (CVPR)}, June 2019.

\bibitem{paszke2017automatic}
Adam Paszke, Sam Gross, Soumith Chintala, Gregory Chanan, Edward Yang, Zachary
  DeVito, Zeming Lin, Alban Desmaison, Luca Antiga, and Adam Lerer.
\newblock Automatic differentiation in pytorch.
\newblock 2017.

\bibitem{qi2017pointnet}
Charles~R Qi, Hao Su, Kaichun Mo, and Leonidas~J Guibas.
\newblock Pointnet: Deep learning on point sets for 3d classification and
  segmentation.
\newblock In {\em Proceedings of the IEEE Conference on Computer Vision and
  Pattern Recognition}, pages 652--660, 2017.

\bibitem{richter2018matryoshka}
Stephan~R Richter and Stefan Roth.
\newblock Matryoshka networks: Predicting 3d geometry via nested shape layers.
\newblock In {\em Proceedings of the IEEE conference on computer vision and
  pattern recognition}, pages 1936--1944, 2018.

\bibitem{sitzmann2020implicit}
Vincent Sitzmann, Julien Martel, Alexander Bergman, David Lindell, and Gordon
  Wetzstein.
\newblock Implicit neural representations with periodic activation functions.
\newblock {\em Advances in Neural Information Processing Systems}, 33, 2020.

\bibitem{slavcheva2017killingfusion}
Miroslava Slavcheva, Maximilian Baust, Daniel Cremers, and Slobodan Ilic.
\newblock Killingfusion: Non-rigid 3d reconstruction without correspondences.
\newblock In {\em Proceedings of the IEEE Conference on Computer Vision and
  Pattern Recognition}, pages 1386--1395, 2017.

\bibitem{solomon11as-killing-as-possible}
Justin Solomon, Mirela Ben{-}Chen, Adrian Butscher, and Leonidas~J. Guibas.
\newblock As-killing-as-possible vector fields for planar deformation.
\newblock {\em Comput. Graph. Forum}, 30(5), 2011.

\bibitem{sorkine2007rigid}
Olga Sorkine and Marc Alexa.
\newblock As-rigid-as-possible surface modeling.
\newblock In {\em Symposium on Geometry processing}, volume~4, pages 109--116,
  2007.

\bibitem{stam2011velocity}
Jos Stam and Ryan Schmidt.
\newblock On the velocity of an implicit surface.
\newblock {\em ACM Transactions on Graphics (TOG)}, 30(3):1--7, 2011.

\bibitem{sun2009concise}
Jian Sun, Maks Ovsjanikov, and Leonidas Guibas.
\newblock A concise and provably informative multi-scale signature based on
  heat diffusion.
\newblock In {\em Computer graphics forum}, volume~28, pages 1383--1392. Wiley
  Online Library, 2009.

\bibitem{tao2016near}
Michael Tao, Justin Solomon, and Adrian Butscher.
\newblock Near-isometric level set tracking.
\newblock In {\em Computer Graphics Forum}, volume~35, pages 65--77. Wiley
  Online Library, 2016.

\bibitem{tatarchenko2016multi}
Maxim Tatarchenko, Alexey Dosovitskiy, and Thomas Brox.
\newblock Multi-view 3d models from single images with a convolutional network.
\newblock In {\em European Conference on Computer Vision}, pages 322--337.
  Springer, 2016.

\bibitem{tatarchenko2017octree}
Maxim Tatarchenko, Alexey Dosovitskiy, and Thomas Brox.
\newblock Octree generating networks: Efficient convolutional architectures for
  high-resolution 3d outputs.
\newblock In {\em Proceedings of the IEEE International Conference on Computer
  Vision}, pages 2088--2096, 2017.

\bibitem{wang2018pixel2mesh}
Nanyang Wang, Yinda Zhang, Zhuwen Li, Yanwei Fu, Wei Liu, and Yu-Gang Jiang.
\newblock Pixel2mesh: Generating 3d mesh models from single rgb images.
\newblock In {\em Proceedings of the European Conference on Computer Vision
  (ECCV)}, pages 52--67, 2018.

\bibitem{wang20193dn}
Weiyue Wang, Duygu Ceylan, Radomir Mech, and Ulrich Neumann.
\newblock 3dn: 3d deformation network.
\newblock In {\em Proceedings of the IEEE Conference on Computer Vision and
  Pattern Recognition}, pages 1038--1046, 2019.

\bibitem{williams2020voronoinet}
Francis Williams, Jerome Parent-Levesque, Derek Nowrouzezahrai, Daniele
  Panozzo, Kwang Moo~Yi, and Andrea Tagliasacchi.
\newblock Voronoinet: General functional approximators with local support.
\newblock In {\em Proceedings of the IEEE/CVF Conference on Computer Vision and
  Pattern Recognition Workshops}, pages 264--265, 2020.

\bibitem{williams2019deep}
Francis Williams, Teseo Schneider, Claudio Silva, Denis Zorin, Joan Bruna, and
  Daniele Panozzo.
\newblock Deep geometric prior for surface reconstruction.
\newblock In {\em Proceedings of the IEEE Conference on Computer Vision and
  Pattern Recognition}, pages 10130--10139, 2019.

\bibitem{wu2016learning}
Jiajun Wu, Chengkai Zhang, Tianfan Xue, Bill Freeman, and Josh Tenenbaum.
\newblock Learning a probabilistic latent space of object shapes via 3d
  generative-adversarial modeling.
\newblock In {\em Advances in neural information processing systems}, pages
  82--90, 2016.

\bibitem{xu2019disn}
Qiangeng Xu, Weiyue Wang, Duygu Ceylan, Radomir Mech, and Ulrich Neumann.
\newblock Disn: Deep implicit surface network for high-quality single-view 3d
  reconstruction.
\newblock In {\em Advances in Neural Information Processing Systems}, pages
  492--502, 2019.

\end{thebibliography}

\clearpage
\section{Appendix}
\begin{figure*}[t]
% \newcommand{\methodheader}[1]{\parbox{3.5cm}{\centering#1}}
% \rotatebox{-90}{%
% \methodheader{Ours}%
% \methodheader{AD}%
% \methodheader{VAE}%

% \methodheader{Ours}%
% \methodheader{AD}%
% \methodheader{VAE}%
% }
\centering

\includegraphics[width=0.96\textwidth]{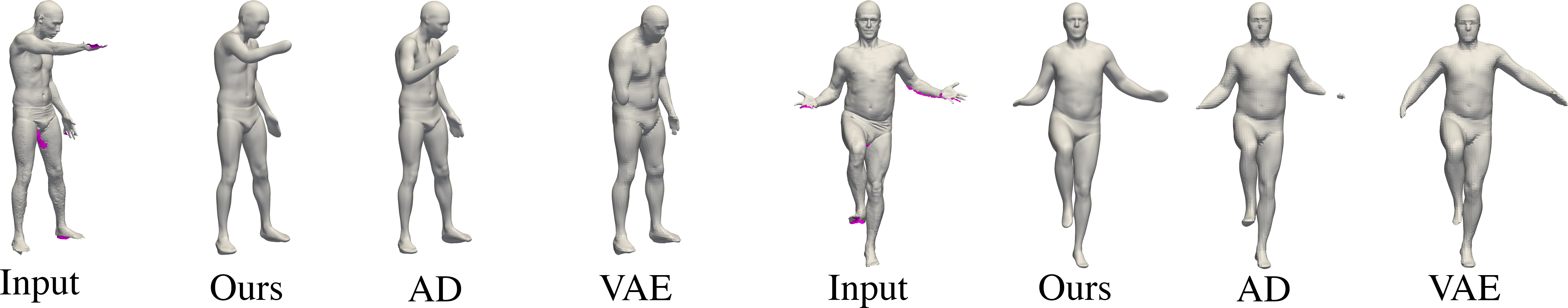} 

\caption{Failures in reconstructing unseen poses. From left to right (in each sequence of 4): unseen raw scan, ours, AD, VAE.} 
\label{fig:fail_test}
  
\end{figure*}

\begin{figure*}[t]
% \newcommand{\methodheader}[1]{\parbox{3.5cm}{\centering#1}}
% \rotatebox{-90}{%
% \methodheader{Ours}%
% \methodheader{AD}%
% \methodheader{VAE}%

% \methodheader{Ours}%
% \methodheader{AD}%
% \methodheader{VAE}%
% }
\centering
\begin{tabular}{l | l }
\includegraphics[width=0.44\textwidth]{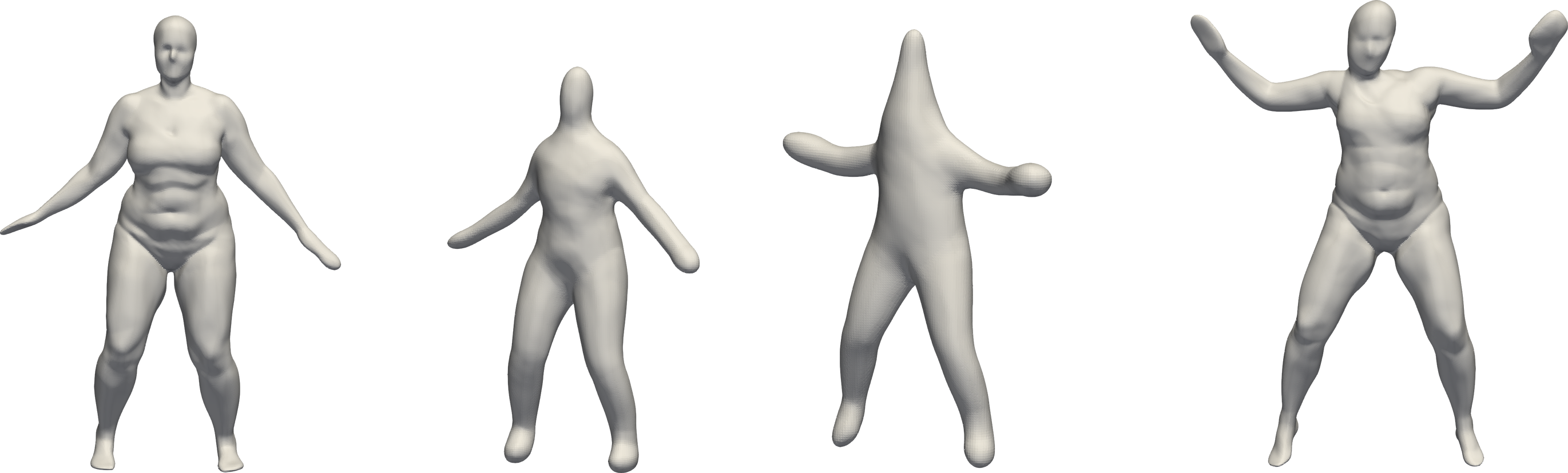}  & \includegraphics[width=0.44\textwidth]{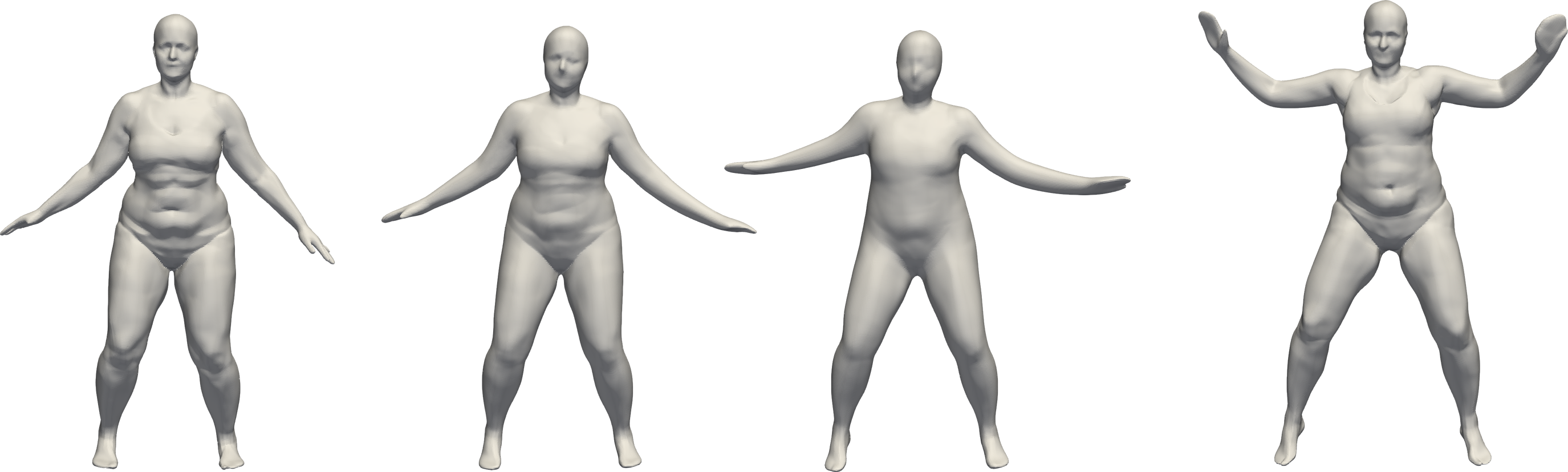}   \\

\multicolumn{1}{c|}{$k=1$} & \multicolumn{1}{c}{$k=5$}  \\
\hline
\includegraphics[width=0.44\textwidth]{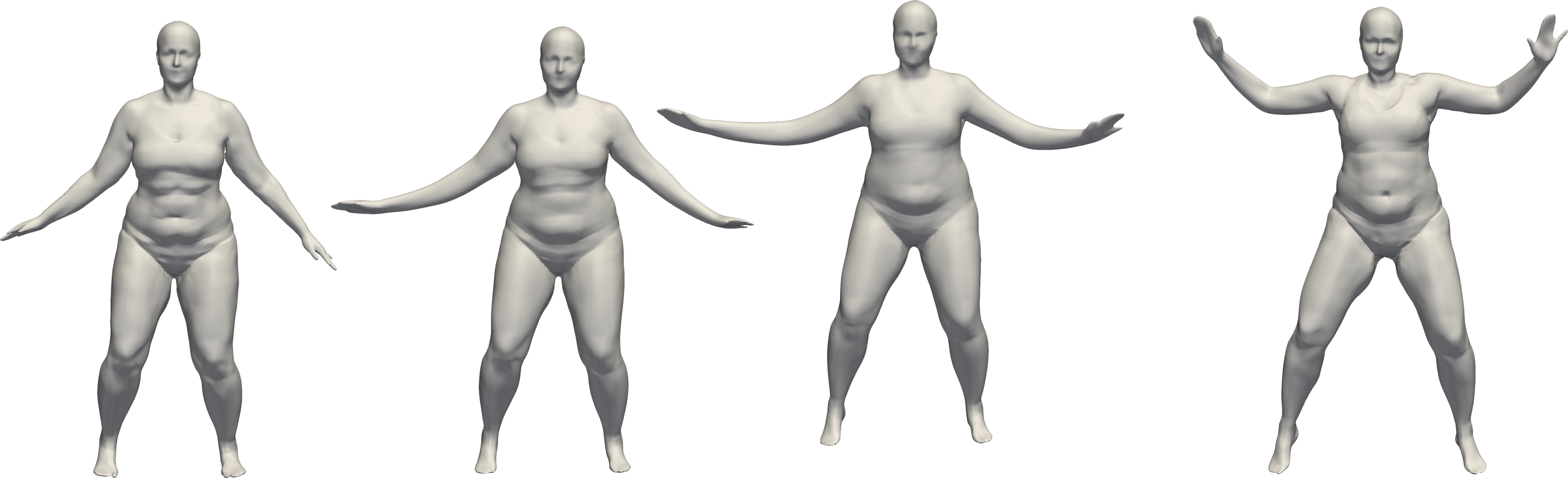}  & \includegraphics[width=0.44\textwidth]{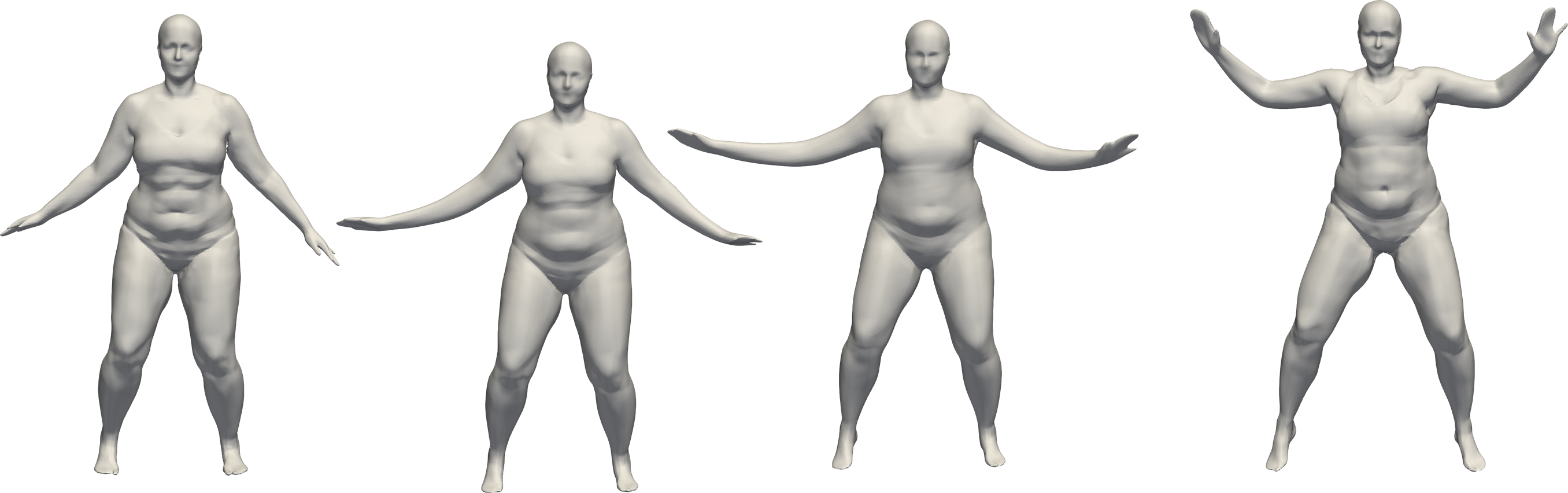}   \\
\multicolumn{1}{c|}{$k=10$} & \multicolumn{1}{c}{$k=15$}\\
\hline
\includegraphics[width=0.44\textwidth]{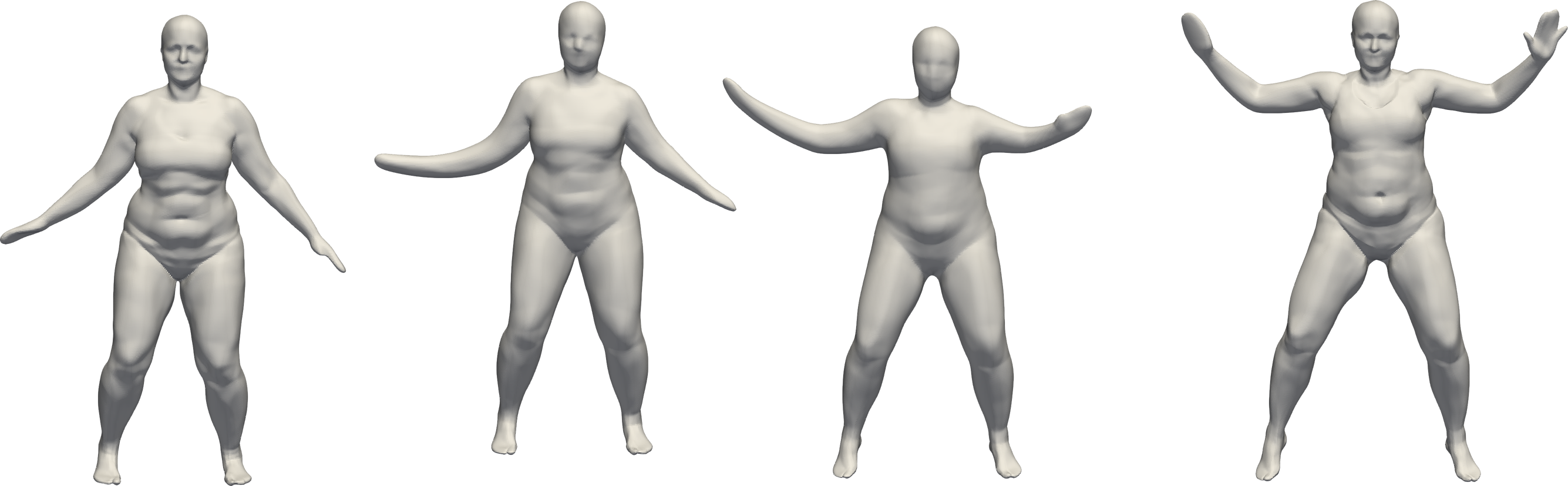}  & \includegraphics[width=0.44\textwidth]{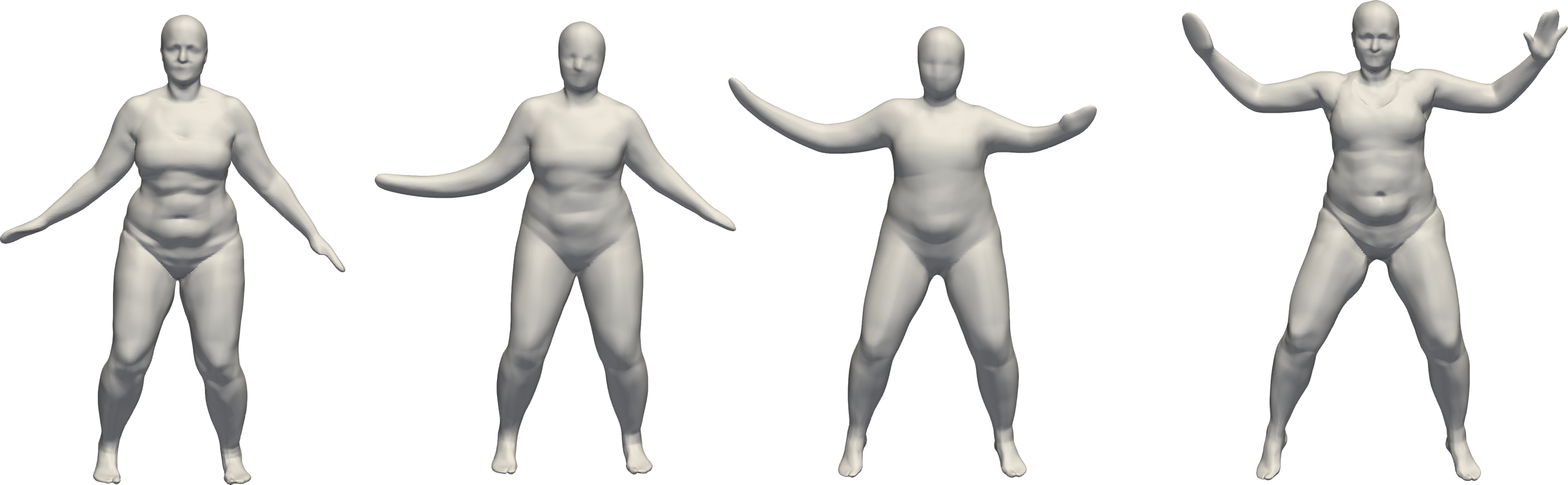}   \\
\multicolumn{1}{c|}{$k=40$} & \multicolumn{1}{c}{$k=200$}\\
\hline
\includegraphics[width=0.44\textwidth]{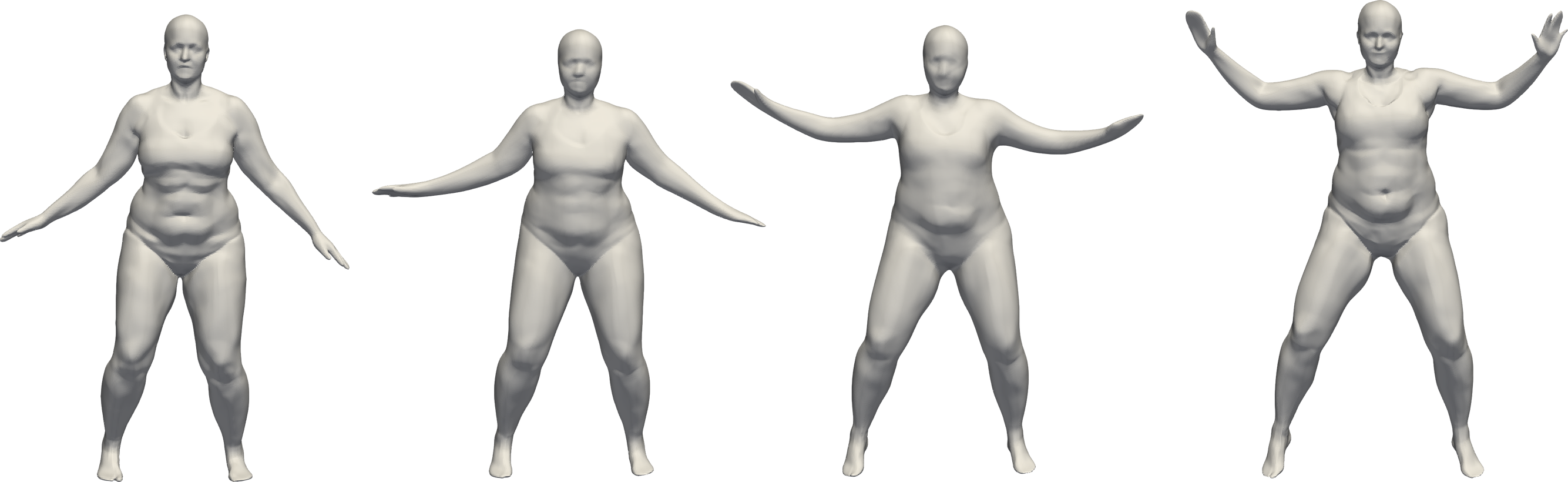}  & \includegraphics[width=0.44\textwidth]{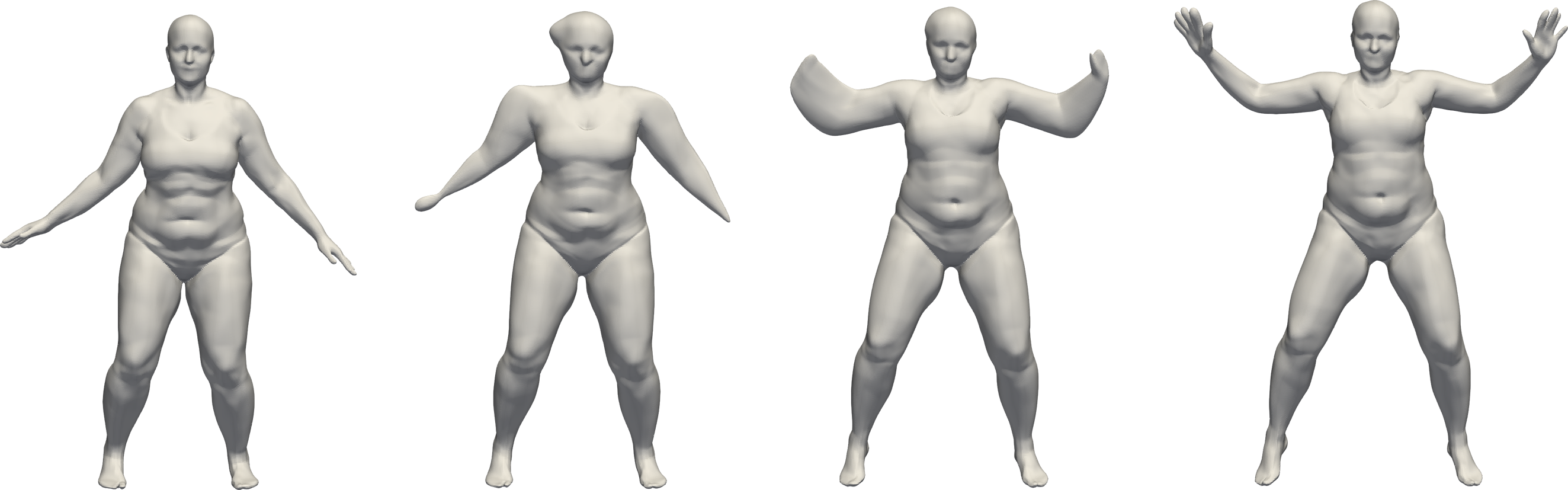}   \\
\multicolumn{1}{c|}{$k=400$} & \multicolumn{1}{c}{$k = \abs{\gX}$}\\
\end{tabular}
\caption{Ablation: experimenting with various $k$ values, the number of affine fields. Left and right in each row are the learned reconstructions. The middle shapes were generated by latent interpolation.} 
\label{tab:ablation}
  
\end{figure*}

\subsection{Additional Results}
\begin{figure*}[t]
% \newcommand{\methodheader}[1]{\parbox{3.5cm}{\centering#1}}
% \rotatebox{-90}{%
% \methodheader{Ours}%
% \methodheader{AD}%
% \methodheader{VAE}%

% \methodheader{Ours}%
% \methodheader{AD}%
% \methodheader{VAE}%
% }
\centering
   
\includegraphics[width=0.96\textwidth]{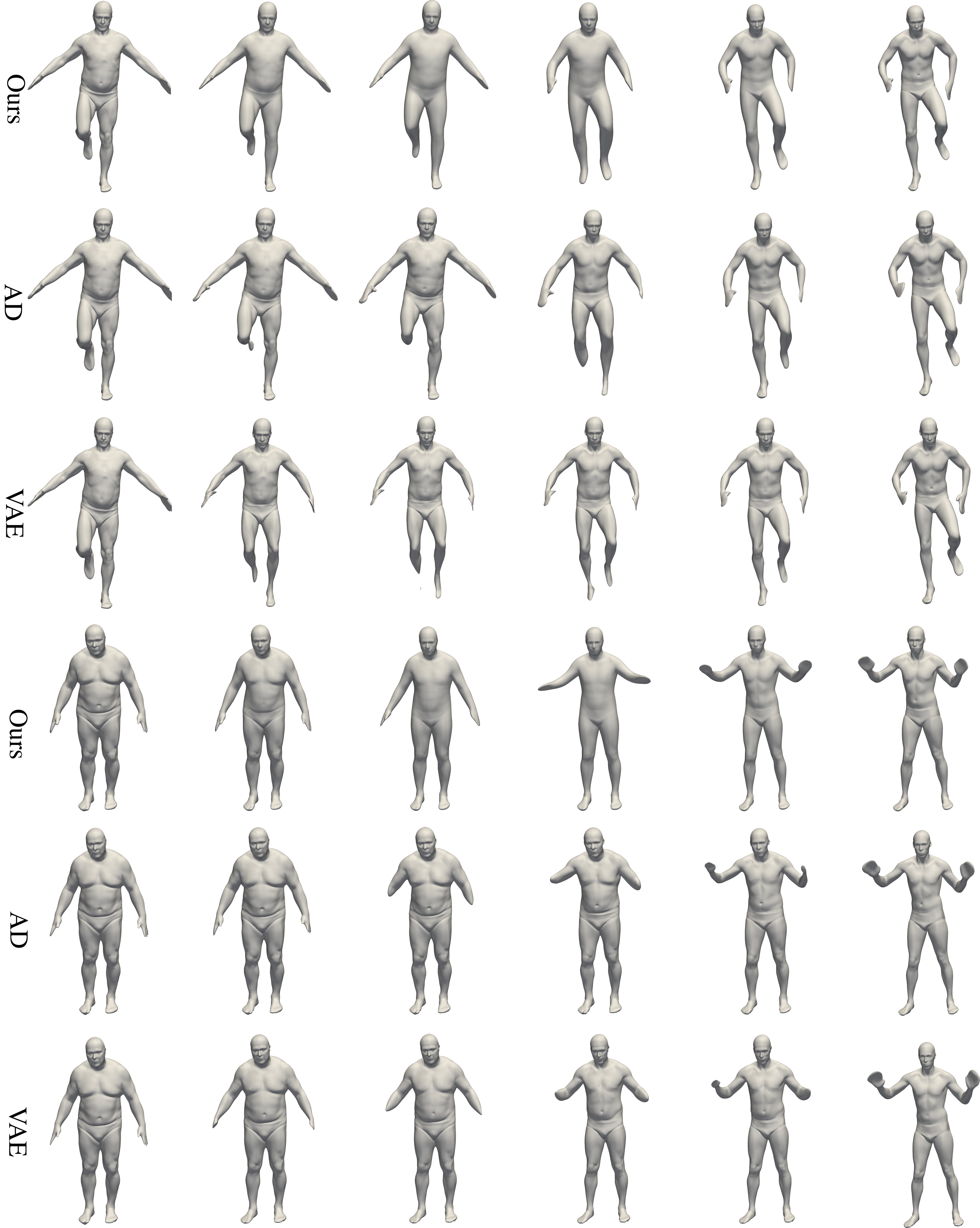} 
            
             \caption{Interpolation of latent codes between diffent humans. Each block of three rows shows (top to bottom): Our result, AD, and VAE.}%\vspace{-15pt}
            \label{fig:large_interpolation}
\end{figure*}

We provide a video file named toy\_figure.mp4 showing animation sequences of results and data used for the evaluation in Section \ref{s:evaluation}. In addition, we provide video files male.mp4 and female.mp4, showing animation sequences of learned interpolations from split 3 in the experiment described in Section \ref{s:shape_space}. For splits 1 and 2, qualitative examples of interpolations between different humans are shown in Figure \ref{fig:large_interpolation}.

\subsection{Ablation over $k$, the number of affine fields}
The main parameter of our method is $k$, the number of affine fields (see \eqref{e:loss_deform}) in our multiple deformation prior loss. The intuition in setting $k$ is that it should capture the number of different parts of a shape that move rigidly. To test the effect of $k$ we designed the following experiment. We considered two input shapes $\gX_1,\gX_2$ and trained using our loss as in \eqref{e:loss_complete}, under different number values for $k$. In each run, all the other hyper-parameters were set to be the same: $\lambda_\text{e}=0.1$, $\lambda_\text{ad}=0.001$, $\lambda_\text{d} = 0.001$ and $D = 1$. Figure \ref{tab:ablation} shows learned input reconstructions (far left and right); and in addition, it shows shapes generated by linear interpolation in the one dimensional latent space. Our findings can be summarized as follows: the number $k$ serves merely as an upper bound to the number of deforming rigid parts that can be explained by the deformation prior loss. That is, $k$ should not be set too small with respect to the expected number of independent deforming rigid parts, but other than that there is no harm in setting $k$ to be much bigger. In figure \ref{tab:ablation} we see comparable results in quality for $k \in \left\{5,10,15,40,200,400\right\}$. Note that setting $k$ too small, as in $k=1$, result in poor quality reconstructions. One question that might raise from these results is why should we limit the number of affine fields, $k$, at all? That is, setting $k=\abs{\gX}$ by allowing a separate rigid deformation prior for each $x\in \gX$. We see that for $k=\abs{\gX}$, we get high quality reconstruction as well. Nonetheless, deformations in latent space are no longer seem to be natural and fail to capture multiple parts rigid deformation between the two input shapes. This may be attributed to the fact that using a different, non-consistent affine field at every point allows too much flexibility and can reduce the killing loss without resulting in a near-rigid deformation.

\subsection{Failure cases}
Figure \ref{fig:fail_test} shows some typical failures of our method. Left part depicts a test case from the removed punching sequence in split 2, whereas the right part is from the removed\_one\_leg\_jump sequence in split 1. In essence, although we demonstrate improvement over baselines there are still cases where the network fails to generalize to unseen poses. See the bend in the right arm in the left example and the missing parts from right hand in the right example. One possible solution could be to train our loss not only between pairs of latents in $\gZ$, but also to more general latent samples in $\Real^D$ to allow better extrapolation. 

\subsection{Additional Implementation details}
\begin{figure}[t]
    \centering
    
         \includegraphics[width=0.96\columnwidth]{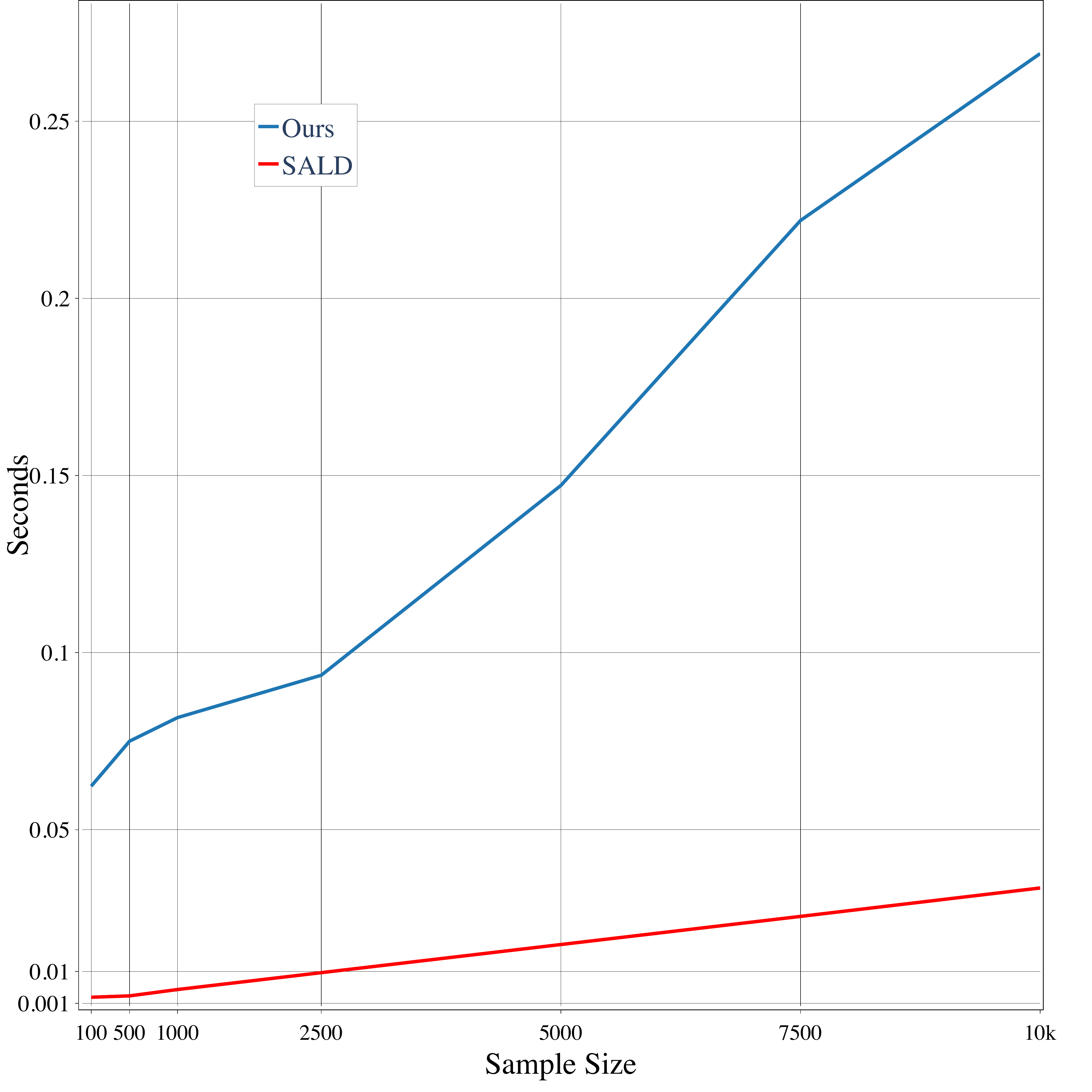} 

    \caption{Timings. We report the time of a single forward pass (in seconds) as a function of sample size.}
    \label{fig:timing}
\end{figure}

\paragraph{Architecture.}
Our architecture consists of two networks $f:\Real^{3}\times\Real^D\too \Real$, and the probability network $\vp=(p_1,\ldots,p_k):\Real^{3}\times\Real^D\too \Real^k$. In all experiements in the paper we used $D=256, k=20$, unless stated otherwise. The network $\vp$ is implemented by a ReLU activation one-layer MLP with 128 hidden units. The network $f$ is implemented by a 8-layers MLP, with a single skip connection between the input to the middle layer. There are $512$ units in each hidden layer. Note that the same architecture was used in \cite{Park_2019_CVPR,atzmon2019sal,gropp2020implicit,atzmon2020sald}. For the initialization of $f$ weights, we use the geometric initialization from \cite{atzmon2019sal}. We use the SoftPlus function, with $\beta=100$ for activation. The same architecture is used for the AutoDecoder baseline and for the decoder in the VAE baseline. For the encoder in the VAE baseline we use the same one as in \cite{atzmon2020sald}. 

Next, we describe additional details regarding the terms that consists our loss, see \eqref{e:loss_complete} in the main paper.

\paragraph{Eikonal loss.}
We utilize the Eikonal loss \cite{gropp2020implicit} to regularize the level-sets of implicit surfaces at intermediate latent codes $\vz_i$
\begin{equation}\label{e:eikonal}
    \loss_{\text{e}}(\theta) = \frac{1}{n}\sum_{i=1}^n \parr{\norm{\nabla_\vx f(\vy_i,\vz_i)}-1}^2,
\end{equation}
where $\vy_i\in\Real^3$ are sampled uniformly in a bounding box of all the input shapes.

\paragraph{Deformation loss.}
Here we give the missing details from the main text regarding the sampling of the points $\vx_i \in \gS(\vz_i)$. We start by drawing a uniform sample of points $\vp_i$ in a bounding box of the shape $\gS(\vz_i)$. Then, we perform $5$ iterations of the generalized Newton, defined by:
\begin{equation}
    \vp_i^{\text{next}} = \vp_i - \frac{\nabla_{\vx} f^T}{\norm{\nabla_{\vx}f }^2} f(\vp_i).
\end{equation}
Finally, we set $\vx_i = \vp_i^{\text{next}} + \vn_i$, where $\vn_i$ is a random sample from a Gaussian noise $\gN(0,0.02)$. 

\paragraph{Reconstruction loss.}

In order to approximate the input shapes $\gX^{(i)}$ at the latent codes $\vz^{(i)}$, we use the SALD reconstruction loss \cite{atzmon2020sald} that handles raw data as input, and only requires the \emph{unsigned} distance to the input geometry, $d(\vx,\gX) = \min_{\vy\in\gX} \norm{\vx-\vy}$.
For a batch of size $b$ of samples $\vq_i, \vz^{(i)}$, $i\in[b]$, the loss is defined by
\begin{align*}
    \loss_{\text{r}}(\theta) &= \frac{1}{b} \sum_{i=1}^b \Big [ \tau\big(f(\vq_i,\vz^{(i)}),d(\vq_i,\gX^{(i)})\big) \\  &+ \lambda \tau\big(\nabla_\vx f(\vq_i, \vz^{(i)}), \nabla_\vx d(\vq_i,\gX^{(i)})\big)\Big ],
\end{align*}
where $\tau$ is the sign agnostic function, \ie, $\tau(a,b) = \min\set{|a-b|,|a+b|}$ for scalars, and $\tau(\va,\vb)=\min\set{\norm{\va-\vb},\norm{\va+\vb}}$ for vectors.
The latent $\vz^{(i)}$ is a random sample from $\gZ$, and the point $\vq_i$ is sampled by adding random displacement, $\vn_i$, to a uniform sample from the input geometry $\vx_i \in \gX^{(i)}$. That is, $\vq_i = \vx_i + \vn_i$. The displacement distribution is a mixture of two isotropic Gaussians, $\gN(0,\sigma_{i,1}^2 I)$ and $\gN_i(0,\sigma_{i,2}^2 I)$. The parameter $\sigma_{i,1}^2$ depends on each $\vx_i$ and is set to be the distance of the \nth{10} closest point to $\vx_i$, whereas $\sigma_{i,2}$ was set to $0.3$ fixed.
Lastly, in all experiments we set $\lambda=0.1$.

\subsubsection{Training details}\label{appendix:training} 
In all the experiments in section \ref{s:shape_space}, including baselines, training was done using the \textsc{Adam} \cite{kingma2014adam} optimizer, with batch size $b=64$. For the VAE, we set a fixed learning rate of $0.0005$. The same learning rate was set for the AD, except for the learning rate of the latents $\left\{\vz^{(i)}\right\}$, which was set to $0.001$. For splits 1 and 2, networks were trained for $5000$ epochs. The loss parameter $\lambda_\text{d}$ was set according to the following scheduling: $\lambda_\text{d} = 0$ for the first $2000$ epochs, followed by $\lambda_\text{d}=0.001$ for the next $2000$ epochs, and $\lambda_\text{d}=0.0001$ for the last $1000$ epochs.  For splits 3 and 4, training was done for 50k epochs. The scheduling for $\lambda_\text{d}$ was to set $\lambda_\text{d}=0$ for the first 20k epochs, followed by $\lambda_\text{d}=0.001$ for the next 20k epochs, and $\lambda_\text{d}=0.0001$ for the last 10k epochs. Training was done on $4$ Nvidia V-100 GPUs, using \textsc{pytorch} deep learning framework \cite{paszke2017automatic}. For the evaluation in section \ref{s:evaluation}, training was for $5000$ epochs for all experiments, on a single Nvidia V-100 GPU. The batch size was $b=8$. In addition, we set $\lambda_\text{d}=0.001$ fixed.

\paragraph{Timings.} In figure \ref{fig:timing} we report the total seconds required for a single forward pass, on a single Nvidia V-100 GPU. The baseline for comparison is SALD \cite{atzmon2020sald}. In the experiments in Section \ref{s:shape_space}, a sample size of $2500$ points was used, resulted in approximately $9.7$ times slower training than SALD. The main additional cost in our method can be attributed to the calculation of the network's second derivatives, as part of the calculation of the rigid deformation prior. 
% \paragraph{Architecture.}

% The implicit network $f$ is implemented as an 8-layers MLP with 512 hidden units in each layer, and a single skip connection between the input to the middle layer.
% This is the same architecture that has been previously used in \cite{Park_2019_CVPR,atzmon2019controlling,atzmon2019sal,gropp2020implicit,atzmon2020sald}.
% For the activation function, we chose SoftPlus with parameter $\beta=100$.
% The MLP $\vp$ is implemented as a ReLU activation One-layer MLP with 128 hidden units in the middle layer. We provide more implementation details in the supplementary.

\subsubsection{Evaluation metrics}
We used two different metrics to measure distances between shapes. One metric is the \emph{Chamfer} distance, $\dist_{\text{C}}$, measuring the squared distance between each point in one shape to its nearest neighbour in the other. Second metric is the \emph{Wasserstein} distance, $\dist_{\text{W}}$, which measures the sum of the distances of the optimal transportation between the shapes.  See \cite{achlioptas2018learning} for more details about these two metrics. Next, we present the formal definition for these metrics. The definition of the \emph{Chamfer} distance is

\begin{align} \label{e:CD}
\dist_{\text{C}}\left(\gX_{1},\gX_{2} \right) & = \frac{1}{2}\left(\dist_{\text{C}}^{\rightarrow}\left(\gX_{1},\gX_{2} \right) + \dist_{\text{C}}^{\rightarrow}\left(\gX_{2},\gX_{1} \right) \right)
\end{align}
where 
\begin{align}\label{e:one_sided_CD}
\dist_{\text{C}}^{\rightarrow}\left(\gX_{1},\gX_{2} \right) & = \frac{1}{\abs{\gX_1}}\sum_{\vx_{1}\in\gX_{1}}\min_{\vx_2\in \gX_{2}}\norm{\vx_1-\vx_2}^2.
\end{align}
The definition for the \emph{Wasserstein} distance is

\begin{align}\label{e:wasser}
\dist_{\text{W}}\left(\gX_{1},\gX_{2} \right) & = \min_{\phi:\gX_1 \rightarrow \gX_2} \sum_{x\in \gX_1}\norm{\phi\left(x\right) - x}
\end{align}
where $\phi$ is a bijection. Note that $\dist_{\text{W}}$ inputs of equal size. In tables \ref{tab:all_humans_splits} and \ref{tab:individual_splits} we evaluated $\dist_{\text{C}}$ on samples of size 30k points, and $\dist_{\text{W}}$ on samples of size 1k points.
\subsubsection{Danskin's Theorem}

\begin{theorem}[Danskin]\label{thm:danskin}
Let $g:\Real\times \Omega \too \Real$ be a continuous function, continuously differentiable in its first argument, where $\Omega\subset \Real^d$ is a compact set.
Let
\begin{equation}\label{e:G}
G(t,\vx) = \min_{\vx\in\Omega}g(t,\vx).
\end{equation}
Then
$$\frac{\partial G}{\partial t}(t,\vx) = \frac{\partial G}{\partial t}(t,\vx_\star),$$
when $\vx_\star$ is the unique minimizer of~\cref{e:G}.
\end{theorem}
This version of Danskin's Theorem follows directly from Corollary 10.1 in~\cite{bacsar2008h}.

\end{document}